\documentclass[lettersize,journal]{IEEEtran}
\usepackage{amsmath,amsfonts} 
\usepackage{algorithm,algcompatible}
\algnewcommand\INPUT{\item[\textbf{Input:}]}
\algnewcommand\OUTPUT{\item[\textbf{Output:}]}%
\usepackage{array}
\usepackage[caption=false,font=normalsize,labelfont=sf,textfont=sf]{subfig}
\usepackage{textcomp}
\usepackage{stfloats}

\usepackage{url}
\usepackage{verbatim}
\usepackage{graphicx}
\usepackage{threeparttable}
\usepackage{soul}

\usepackage{cite}
\usepackage{multirow}
\usepackage{color}
\usepackage[dvipsnames]{xcolor}
\usepackage{tikz}
\DeclareRobustCommand{\blackline}{\raisebox{2pt}{\tikz{\draw[-,black!40!Black,solid,line width = 0.9pt](0,0) -- (5mm,0);}}}
\DeclareRobustCommand{\blueline}{\raisebox{2pt}{\tikz{\draw[RoyalBlue,solid,line width = 0.9pt](0,0) -- (5mm,0);}}}
\DeclareRobustCommand{\orangeline}{\raisebox{2pt}{\tikz{\draw[BurntOrange,solid,line width = 0.9pt](0,0) -- (5mm,0);}}}
\DeclareRobustCommand{\greenline}{\raisebox{2pt}{\tikz{\draw[Green,solid,line width = 0.9pt](0,0) -- (5mm,0);}}}
\DeclareRobustCommand{\redline}{\raisebox{2pt}{\tikz{\draw[Red,solid,line width = 0.9pt](0,0) -- (5mm,0);}}}
\DeclareRobustCommand{\purpleline}{\raisebox{2pt}{\tikz{\draw[Purple,solid,line width = 0.9pt](0,0) -- (5mm,0);}}}
\DeclareRobustCommand{\blackdashedline}{\raisebox{2pt}{\tikz{\draw[-,black!40!black,dashed,line width = 0.9pt](0,0) -- (5mm,0);}}}

\DeclareRobustCommand{\reddashedline}{\raisebox{2pt}{\tikz{\draw[Red,dashed,line width = 0.9pt](0,0) -- (5mm,0);}}}

\DeclareRobustCommand{\tikzcircle}[2][red,fill=red]{\tikz[baseline=-0.5ex]\draw[#1,radius=#2] (0,0) circle ;}
\DeclareRobustCommand{\tikzcirclehollow}[2][red]{\tikz[baseline=-0.5ex]\draw[#1,radius=#2] (0,0) circle;}

\usepackage[switch]{lineno}

\begin{document}

\title{An adaptive multi-fidelity sampling framework for safety analysis of connected and automated vehicles}

\author{Xianliang Gong, Shuo Feng, Yulin Pan
\thanks{Xianliang Gong and Yulin Pan are with the Department of Naval Architecture and Marine Engineering, University of Michigan, 48109, MI, USA (e-mail: xlgong@umich.edu; yulinpan@umich.edu)}
\thanks{Shuo Feng is with the Department of Automation, Tsinghua University, Beijing 100084, China, (e-mail: fshuo@tsinghua.edu.cn)}
}



\maketitle
\begin{abstract}
Testing and evaluation are expensive but critical steps  in the development of connected and automated vehicles (CAVs). In this paper, we develop an adaptive sampling framework to efficiently evaluate the accident rate of CAVs, particularly for scenario-based tests where the probability distribution of input parameters is known from the Naturalistic Driving Data. Our framework relies on a surrogate model to approximate the CAV performance and a novel acquisition function to maximize the benefit (information to accident rate) of the next sample formulated through an information-theoretic consideration. In addition to the standard application with only a single high-fidelity model of CAV performance, we also extend our approach to the bi-fidelity context where an additional low-fidelity model can be used at a lower computational cost to approximate the CAV performance. Accordingly, for the second case, our approach is formulated such that it allows the choice of the next sample in terms of both fidelity level (i.e., which model to use) and sampling location to maximize the benefit per cost. Our framework is tested in a widely-considered two-dimensional cut-in problem for CAVs, where Intelligent Driving Model (IDM) with different time resolutions are used to construct the high and low-fidelity models. We show that our single-fidelity method outperforms the existing approach for the same problem, and the bi-fidelity method can further save half of the computational cost to reach a similar accuracy in estimating the accident rate.

\end{abstract}


\section{Introduction}
\IEEEPARstart{C}{onnected} and autonomous vehicles (CAVs) have attracted increasing attention due to their potential to improve mobility and safety while reducing the energy consumed. One critical issue for the development of CAVs is their safety testing and evaluation. In general, converged statistics of accident rate may require hundreds of millions of miles for each configuration of CAVs \cite{RR-1478-RC}. To reduce the testing cost, scenario-based approaches have been developed \cite{riedmaier2020survey, nalic2020scenario, ulbrich2015defining, sun2021scenario}, with many of them testing certain simplified driving events, e.g., the cut-in problem. In the scenario-based framework, the scenarios (and their distribution) describing certain traffic environment are parameterized from the Naturalistic Driving Data (NDD). The performance of CAVs is then evaluated for given scenarios as the input, and the accident rate of CAVs is quantified considering the distribution of scenarios. 

The scenario-based safety analysis, however, is far from a trivial task. The difficulties lie in the high cost of evaluating the CAV performance given a scenario (say, using road tests or high-fidelity simulators) and the rareness of accidents in the scenario space \cite{liu2022curse}. The two factors result in a large number of required CAV performance evaluations that can become prohibitively expensive (either computationally or financially) if a standard Monte Carlo method is used. In order to address the problem, many methods have been developed to reduce the number of scenario evaluations in safety analysis. One category of methods rely on importance sampling, where samples are selected from a proposal distribution to stress the critical input regions (leading to most accidents). Different ways to construct the proposal distribution have been developed in \cite{zhao2016accelerated, huang2017accelerated, feng2020testing1, feng2021intelligent, feng2020testing3},
leading to significant acceleration compared to the Monte Carlo method. {\color{black} In particular, the proposal distribution is constructed in \cite{zhao2016accelerated} from a parametric distribution with its parameters determined from the cross-entropy method \cite{rubinstein1999cross}. The method in \cite{zhao2016accelerated} is further improved in \cite{huang2017accelerated} by employing piece-wise parametric distributions (i.e., different parameters used for different parts of input space). The proposal distribution can also be constructed via a low-fidelity model (assumed to be associated with negligible cost), either from the low-fidelity model evaluation itself \cite{feng2020testing1,feng2021intelligent} or additionally leveraging a small number of adaptively selected high-fidelity model evaluations \cite{feng2020testing3}.}

%

Another category of methods in safety analysis is based on adaptive sampling enabled by active learning method. Under this approach, a proposal distribution is not needed, and one directly computes the accident rate according to the input scenario probability with a surrogate model approximating the CAV performance. The surrogate model can be established through a supervised learning approach, say a Gaussian process regression, together with an adaptive sampling algorithm to choose the next-best sample through optimization of a pre-defined acquisition function. Such choice of the next sample is expected to accelerate the convergence of the accident rate computed from the updated surrogate. This class of methods were first developed for structural reliability analysis \cite{teixeira2021adaptive, echard2011ak, hu2016global, bichon2008efficient, sun2017lif, wang2016gaussian} and have recently been introduced to the CAV field \cite{mullins2018adaptive, sun2021adaptive, lu4341960adaptive, huang2017towards, feng2020testing3}.

{\color{black} To provide more details, two acquisition functions are proposed in \cite{mullins2018adaptive}, respectively designed for Gaussian process regression and k-nearest neighbors as surrogate models, in order to better resolve performance boundaries between accidents and safe scenarios. These acquisition functions combine exploration and exploitation under some heuristic consideration of the surrogate models. The approach in \cite{mullins2018adaptive} is extended in \cite{lu4341960adaptive} by clustering samples into different groups that allow a parallel search of optimal samples to accelerate the overall algorithm. In \cite{sun2021adaptive}, the authors develop two acquisition functions applicable to six different surrogate models, which favor samples expected to respectively produce (i) poor performance, and (ii) performance close to accident threshold, and in the meanwhile, far from existing samples (for exploration). In these works, the proposed acquisition functions are rather empirical and cannot guarantee optimal convergence of the accident rate. In addition, an acquisition function that directly targets the accident rate (in which sense similar to what we develop) is proposed in \cite{huang2017towards, huang2018synthesis}, but their method is not sufficiently supported by numerical tests provided in their papers. A summary of these existing methods is provided in Table \ref{review} (also see \cite{sun2021scenario} for a more comprehensive review). In viewing the state-of-the-art methods in the field, it is clear that large room exists for further improvement of the sampling efficiency (i.e., reduction of the required number of samples) through a more rigorous information-theoretic approach to develop the acquisition. Such developments are not only desired to reduce the cost of CAV safety evaluation but are also valuable to the general field of reliability analysis.}

\begin{table*}
\begin{center}
\caption{Summary of existing adaptive sampling methods and current work}
\begin{threeparttable}[t]
\begin{tabular}{|c|l|ll|l|}
\hline
\multicolumn{1}{|l|}{} &
  \multicolumn{1}{c|}{Surrogate\tnote{1}} &
  \multicolumn{2}{c|}{Acquisition} &
  \multicolumn{1}{c|}{Objective} \\ \hline
\multirow{2}{*}{\cite{mullins2018adaptive, lu4341960adaptive}} &
  GPR &
  \multicolumn{2}{l|}{\begin{tabular}[c]{@{}l@{}}Heuristic combination of exploitation (large gradient of GPR mean)\\ and exploration (large uncertainty of GPR)\end{tabular}} &
  Performance boundary \\ \cline{2-5} 
 &
  KNN &
  \multicolumn{2}{l|}{\begin{tabular}[c]{@{}l@{}}Heuristic combination of exploitation (large variance among neighboring samples)\\ and exploration (large distance to neighboring samples)\end{tabular}} &
  Performance boundary \\ \hline
\multirow{2}{*}{\cite{sun2021adaptive}} &
  \multirow{2}{*}{\begin{tabular}[c]{@{}l@{}}GPR, KNN, \\ XGB, etc.\end{tabular}} &
  \multicolumn{2}{l|}{\begin{tabular}[c]{@{}l@{}}Heuristic combination of exploitation (poor performance) \\ and exploration (large distance to all existing samples)\end{tabular}} &
  Accident scenarios \\ \cline{3-5} 
 &
   &
  \multicolumn{2}{l|}{\begin{tabular}[c]{@{}l@{}}Heuristic combination of exploitation (performance close to accidents threshold) \\ and exploration (large  distance to all existing samples)\end{tabular}} &
  Performance boundary \\ \hline
\cite{feng2020testing3} &
  GPR &
  \multicolumn{2}{l|}{Reducing variance of the proposal distribution} &
  \begin{tabular}[c]{@{}l@{}}Optimal proposal distribution \\ for importance sampling\end{tabular} \\ \hline
\cite{huang2017towards, huang2018synthesis} &
  GPR &
  \multicolumn{2}{l|}{Squared change of accident rate after adding a hypothetical sample} &
  Accident rate \\ \hline
Current work &
  GPR &
  \multicolumn{2}{l|} 
  {\begin{tabular}[c]{@{}l@{}} Approximation of the expected K-L divergence between current and next-step accident \\ rate distribution after adding a hypothetical sample\end{tabular}} &
  Accident rate \\
  \hline
\end{tabular}
\begin{tablenotes}
     \item[1] GPR KNN, and XGB respectively stand for Gaussian process regression, k-nearest neighbors, and extreme gradient boosting.
\end{tablenotes}
\end{threeparttable}
\label{review}
\end{center}
\end{table*}

The cost in the evaluation of CAV accident rate can also be reduced by leveraging low-fidelity models applied in conjunction with the high-fidelity model. In principle, the low-fidelity models can provide useful information on the surrogate model (e.g., the general trend of the function) although their own predictions may be associated with considerable errors. For example, low-fidelity models have been used to generate the proposal distribution for importance sampling \cite{feng2020testing1, feng2020testing3}. It needs to be emphasized that almost all existing works (in the CAV field) assume that the low-fidelity models are associated with negligible cost, i.e., the low-fidelity map from scenario space to CAV performance can be considered as a known function. However, in practical situations, the cost ratio between high and low-fidelity models may not be that drastic. Typical cases include (i) CARLA \cite{Dosovitskiy17} simulator versus SUMO simulator \cite{SUMO2018}, (ii) the same simulator with fine versus coarse-time resolutions. For these cases, a new adaptive-sampling algorithm considering the cost ratio is needed, which is expected to be able to select both the model (i.e., fidelity level) and scenario for the next-best sample to reduce the overall cost in the evaluation of the accident rate. Such methods are not yet available for CAV testing.

In this work, we develop an adaptive sampling algorithm in the active learning framework for safety testing and evaluations of CAVs. {\color{black} The novelty of our method lies in the development of an information-theoretic-based acquisition function that leads to very high sampling efficiency and can be extended to bi-fidelity contexts in a relatively straightforward manner. In particular, our method is applied to two situations: (i) the single-fidelity context where only a high-fidelity model is available; and (ii) the bi-fidelity context where the high-to-low model cost ratio is finite and fixed.} We note that for case (ii), our method needs to be established by using a bi-fidelity Gaussian process as the surrogate model and an acquisition function to select the next sample (in terms of both model fidelity and traffic scenario) which maximizes information gain per cost. Both applications of our method are tested in a widely-considered two-dimensional cut-in problem for CAVs, with the high-fidelity model taken as the Intelligent Driving Model (IDM) with fine time resolution. The low-fidelity model is constructed by a coarser-time-resolution IDM model in application (ii). We compare the performance of our method with the state-of-the-art approaches in the CAV field for the same problem and find that even the single-fidelity approach can considerably outperform the existing approaches. The method in application (ii) can further reduce the computational cost by at least a factor of 2. 

We finally remark that the method we develop here is new to the entire field of reliability analysis according to our knowledge, and its application to other fields may prove equally fruitful. {\color{black} For example, it may be applied to evaluate the ship capsizing probability in ocean engineering \cite{gong2022efficient, gong2022multi}, structural safety analysis \cite{hu2016global, wang2016gaussian}, probability of extreme pandemic spikes for public health \cite{pickering2022discovering} and many other physical, engineering and societal problems. Within the CAV field, our method is certainly not limited to the IDM models used in this paper as demonstrations. It can be connected to a broad range of CAV evaluation tools across on-road tests, closed-facility tests, simulations based on various kinds of simulators (e.g. Google/Waymo’s Car-Craft9 \cite{kaur2021survey}, Intel’s CARLA6 \cite{dosovitskiy2017carla}, Microsoft’s Air-Sim7 \cite{shah2018airsim}, NVIDIA’s Drive Constellation \cite{NVIDIA}). Among these examples, we would like to emphasize the possible benefit of our method to the augmented-reality test environment combing a real test vehicle on road and simulated background vehicles \cite{feng2023dense}. Due to the bi-fidelity capability of our method, it also becomes beneficial to combine two different tools in the above list to further improve the testing efficiency. }The extension of our method to high-dimensional problems is also possible (see \cite{gong2022multi} for another sampling purpose) but will not be considered in this paper.



The python code for the algorithm, named MFGPreliability, is available on Github\footnote{https://github.com/umbrellagong/MFGPreliability}.  

\begin{table}[h]
\caption{NOTATIONS OF VARIABLES}
\begin{center}
\begin{tabular}{cl}
\hline
\textbf{Variable}       & \multicolumn{1}{c}{\textbf{Notation}}                                   \\ \hline
$\mathbf{x} \sim p_\mathbf{x}(\mathbf{x})$     & Decision variables with its probability distribution \\ \hline
$f_h, f_l$ & \begin{tabular}[c]{@{}l@{}}High and low-fidelity models mapping from decision \\ variables to  a measure of CAV performance\end{tabular} \\ \hline
$R_0, \dot{R}_0$ & Range and range rate at the cut-in moment \\ \hline
$P_a$            & Accident rate                             \\ \hline
$\delta$         & Threshold to define an accident           \\ \hline
$\mathcal{D} = \{\mathbf{X}, \mathbf{Y}\}$    & 
\begin{tabular}[c]{@{}l@{}}Existing dataset with inputs $\mathbf{X}$ and corresponding  \\ outputs $\mathbf{Y}$ \end{tabular}
\\ \hline
$k, \theta$      & Kernel function and its hyperparameter     \\ \hline
$U$              & An upper bound of the uncertainty in estimating  $P_a$     \\ \hline
$B$              & Benefit of adding a sequential sample      \\ \hline
$c_h, c_l$       & Cost of high and low-fidelity models      \\ \hline
$\Delta t$       & Time step to simulate the IDM model       \\ \hline

\end{tabular}
\end{center}
\end{table}

\section{Problem setup}  
\begin{figure}
    \centering
    \includegraphics[width=5.5cm]{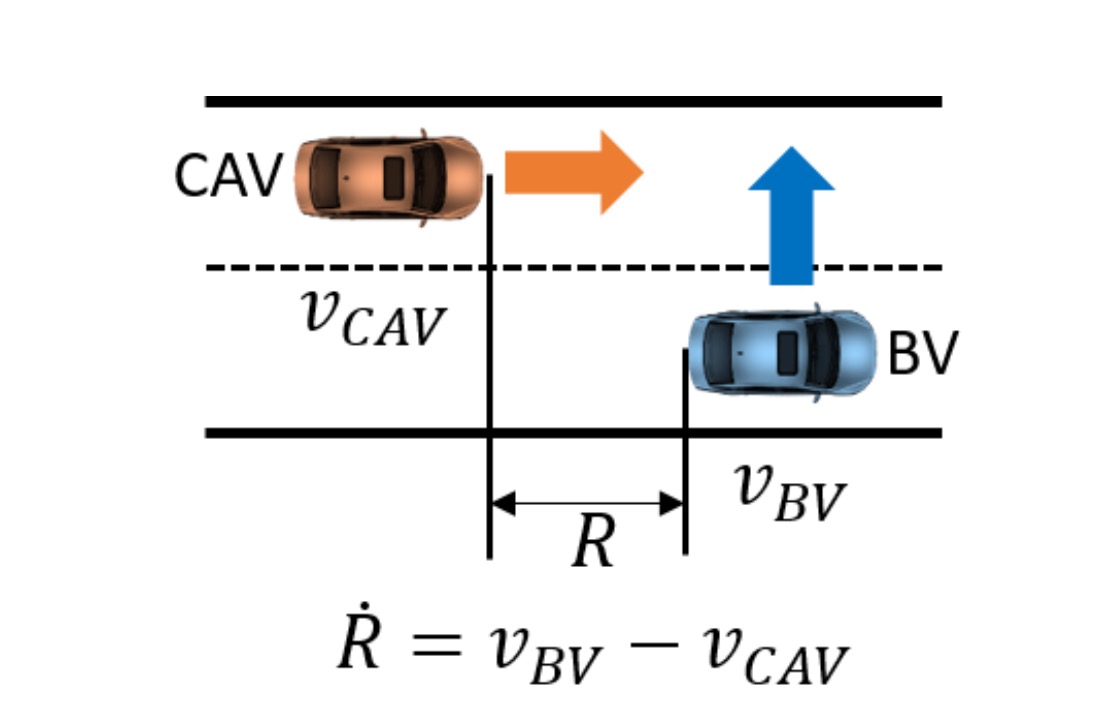}
    \caption{Illustration of the cut-in scenario \cite{feng2020testing3}. $R$ and $\dot{R}$ respectively denote the range and range rate between CAV and BV.}
    \label{fig:cut-in}
\end{figure}

We consider a black-box function $f_h(\mathbf{x}): \mathbb{R}^{d} \to \mathbb{R}$ with input $\mathbf{x}$ a $d$-dimensional decision variable of a driving scenario and output a measure of the CAV performance. A subscript $h$ is used here to denote that the function needs to be evaluated by an expensive high-fidelity model. Taking the cut-in problem (Fig. \ref{fig:cut-in}) as an example, the input can be formulated as $\mathbf{x} = (R_0, \dot{R}_0)$ where $R_0$ and $\dot{R}_0$ denote the initial range and range rate between the CAV and background vehicle (BV) at the cut-in moment $t=0$ (more details in Sec. \ref{case}). The output is the minimum range between the two vehicles during their speed adjustment process for $t\geq 0$.  

The probability of the input $\mathbf{x} \sim p_\mathbf{x}(\mathbf{x})$ is assumed to be known from the naturalistic driving data (NDD). Our objective is the evaluation of accident rate, i.e., probability of the output smaller than some threshold $\delta$ (or range between CAV and BV smaller than $\delta$):
\begin{equation}
    P_a = \int \mathbf{1}_{\delta}(f_h(\mathbf{x}))p_{\mathbf{x}}(\mathbf{x}){\rm{d}}\mathbf{x},
\label{FP}
\end{equation}
where
\begin{equation}
\mathbf{1}_{\delta}(f_h(\mathbf{x})) = \left\{
\begin{aligned}
    & 1,         & 
    & {\rm{if}} \; f_h(\mathbf{x}) < \delta   \\
    & 0,                              & 
    & {\rm{o.w.}} 
\end{aligned}
\right..
\label{indicator}
\end{equation}
A brute-force computation of $P_a$ calls for a large number of Monte Carlo samples in the space of $\mathbf{x}$, which may become computationally prohibitive (considering the expensive evaluation of $f_h$ and the small $P_a$). In this work, we seek to develop an adaptive sampling framework based on active learning, where samples are selected optimally to accelerate the convergence of the computed value of $P_a$. We will present algorithms for (i) single-fidelity cases, where only one model $f_h$ is available, and (ii) bi-fidelity cases. For case (ii), we consider a practical situation that a low-fidelity model $f_l$ with a lower but finite cost is also available to us that can provide a certain level of approximation to $f_h$. Making use of $f_l$, as will be demonstrated, can further reduce the cost of computing $P_a$.

\section{Method}
\subsection{Single fidelity method}
We consider the single-fidelity context where only the model $f_h$ is available. Two basic components of our active learning method are presented below: (i) an inexpensive surrogate model based on the standard Gaussian process; (ii) a new acquisitive function to select the next-best sample.

\subsubsection{surrogate model by GPR}
Gaussian process regression (GPR) is a probabilistic machine learning approach \cite{rasmussen2003gaussian} widely used for active learning. Consider the task of inferring $f_h$ from $\mathcal{D}=\{\mathbf{X}, \mathbf{Y}\}$, which consists of $n$ inputs $\mathbf{X} = \{\mathbf{x}^{i} \in \mathbb{R}^d\}_{i=1}^{i=n}$ and the corresponding outputs $\mathbf{Y} = \{f_h(\mathbf{x}^i)\in \mathbb{R}\}_{i=1}^{i=n}$. In GPR, a prior, representing our beliefs over all possible functions we expect to observe, is placed on $f_h$ as a Gaussian process $f_h(\mathbf{x}) \sim \mathcal{GP}(0,k(\mathbf{x},\mathbf{x}'))$ with zero mean and covariance function $k$ (usually defined by a radial-basis-function kernel): 
\begin{equation}
    k(\mathbf{x},\mathbf{x}') = \tau^2 {\rm{exp}}(-\frac{1}{2} \sum_{j=1}^{j=d}\frac{(x_j-x_j')^2}{s_j^2} ), 
\label{RBF}
\end{equation}
where the amplitude $\tau^2$ and length scales $s_j$ are hyperparameters $\mathbf{\theta}=\{\tau, s_j\}$. 

Following the Bayes' theorem, the posterior prediction for $f_h$ given the dataset $\mathcal{D}$ can be derived to be another Gaussian: 
\begin{linenomath} \begin{align}
    p(f_h(\mathbf{x})|\mathcal{D}) & = \frac{p(f_h(\mathbf{x}),\mathbf{Y})}{p(\mathbf{Y})} 
\nonumber \\
    & = \mathcal{N}\big(\mathbb{E}(f_h(\mathbf{x})|\mathcal{D}), {\rm{cov}}(f_h(\mathbf{x}), f_h(\mathbf{x}')|\mathcal{D})\big),
\label{sgp1}
\end{align} \end{linenomath} 
with mean and covariance respectively:
\begin{IEEEeqnarray}{rCl}
     \mathbb{E}(f_h(\mathbf{x})|\mathcal{D}) &=&  k(\mathbf{x}, \mathbf{X}) {\rm{K}}(\mathbf{X},\mathbf{X})^{-1} \mathbf{Y}, 
\label{sgp2} \\
     {\rm{cov}}(f_h(\mathbf{x}), f_h(\mathbf{x}')|\mathcal{D}) &=& k(\mathbf{x},\mathbf{x}') 
\nonumber \\   
     && - \; k(\mathbf{x},\mathbf{X}) {\rm{K}}(\mathbf{X},\mathbf{X})^{-1} k(\mathbf{X}, \mathbf{x}'), \IEEEeqnarraynumspace
\label{sgp3}
\end{IEEEeqnarray}
where matrix element ${\rm{K}}(\mathbf{X},\mathbf{X})_{ij}=k(\mathbf{x}^i,\mathbf{x}^j)$. The hyperparameters $\mathbf{\theta}$ are determined to be those which maximize the likelihood function $p(\mathcal{D}|\mathbf{\theta})\equiv p(\mathbf{Y}|\mathbf{\theta})=\mathcal{N}(\mathbf{Y};0, {\rm{K}}(\mathbf{X},\mathbf{X}))$. 

\subsubsection{acquisition function}
Given the GPR surrogate $f_h(\mathbf{x})|\mathcal{D}$, the accident rate $P_a|\mathcal{D}$ becomes a random variable with its randomness coming from the uncertainty of the GPR. The principle of finding the next-best sample is to provide most information to the quantity of interest $P_a$. This can be achieved in two ways: (i) through an information-theoretic perspective for the next sample to maximize the information gain, i.e. the K-L divergence between the current estimation and the hypothetical next-step estimation of $P_a$; (ii) through a more intuitive and efficient approach for the next sample to minimize the uncertainty level associated with the distribution of $P_a$. In this paper, we describe the algorithm for (ii) in the main text and the algorithm for (i) in Appendix A. The results from the two approaches are equivalent after simplification of the results from (i) under reasonable assumptions presented in Appendix A. 

For approach (ii), we need to formulate the uncertainty of $P_a$ (measured by the variance of its distribution) after adding a hypothetical sample at $\tilde{\mathbf{x}}$:
\small
\begin{equation}
    \mathrm{var}(P_a|\mathcal{D},\tilde{\mathbf{z}}) 
    =
    \mathrm{var}\Big(\int \mathbf{1}_{\delta}\big(f_h(\mathbf{x})|\mathcal{D},\tilde{\mathbf{z}}\big) p_{\mathbf{x}}(\mathbf{x}){\rm{d}}\mathbf{x}\Big),
\label{ivr_cov} 
\end{equation}
\normalsize
where $\tilde{\mathbf{z}} = \{\tilde{\mathbf{x}}, \overline{f}_h(\tilde{\mathbf{x}})\}$ a new data point with  $\overline{f}_h(\tilde{\mathbf{x}}) = \mathbb{E}(f_h(\tilde{\mathbf{x}})|\mathcal{D})$ the CAV performance computed as the mean prediction from the current GPR. Our purpose is to find $\tilde{\mathbf{x}}$ so that \eqref{ivr_cov} is minimized. However, the computation of \eqref{ivr_cov} is very expensive since the variance operator involves the sampling of an integral (i.e., integral to be computed many times with expensive sampling of $f_h(\mathbf{x})|\mathcal{D}, \tilde{\mathbf{z}}$). The computational cost of \eqref{ivr_cov} can be significantly reduced by considering an upper bound of \eqref{ivr_cov}, following approaches developed in \cite{gong2022sequential}. With the detailed derivation presented in Appendix B, the upper bound of \eqref{ivr_cov} gives 
\begin{equation}
    U(\mathcal{D}, \tilde{\mathbf{z}}) =\int \mathrm{var}^{\frac{1}{2}} \Big(\mathbf{1}_{\delta}\big(f_h(\mathbf{x})|\mathcal{D}, \tilde{\mathbf{z}}\big)\Big) p_{\mathbf{x}}(\mathbf{x}){\rm{d}}\mathbf{x},
\label{ivr}
\end{equation}
where the variance function in \eqref{ivr} can be analytically evaluated (since the indicator function simply follows a Bernoulli distribution for each $\mathbf{x}$) as 
\small
\begin{linenomath} \begin{align}
     \mathrm{var}\Big(\mathbf{1}_{\delta}\big(f_h(\mathbf{x})|\mathcal{D},\tilde{\mathbf{z}}\big)\Big) 
     = & \; \Big(1 -  \Phi\big(\frac{\mathbb{E}\big(f_h(\mathbf{x})|\mathcal{D}, \tilde{\mathbf{z}}\big) - \delta}{\mathrm{var}^{\frac{1}{2}}\big(f_h(\mathbf{x})|\mathcal{D}, \tilde{\mathbf{z}}\big)}\big)\Big) \;
\nonumber \\
      & *   \Phi\big(\frac{\mathbb{E}\big(f_h(\mathbf{x})|\mathcal{D}, \tilde{\mathbf{z}}\big) - \delta}{\mathrm{var}^{\frac{1}{2}}\big(f_h(\mathbf{x})|\mathcal{D}, \tilde{\mathbf{z}}\big)}\big),
\label{var}
\end{align} \end{linenomath} 
\normalsize
with $\Phi$ the cumulative distribution function of a standard Gaussian. It is clear that in evaluating \eqref{var}, no sampling for $f_h(\mathbf{x})|\mathcal{D},\tilde{\mathbf{z}}$ is needed and the integration only needs to be evaluated once, leading to a much cheaper computation compared to \eqref{ivr_cov}. Furthermore, while \eqref{var} seems to involve an updated GPR conditioning on $\{\mathcal{D},\tilde{\mathbf{z}}\}$, the relevant quantities can be efficiently computed using the currently available GPR conditioning on $\mathcal{D}$  (i.e., no update on GPR is needed, see \cite{gong2022multi} for derivation):
\small
\begin{linenomath} \begin{align}
   \mathbb{E}(f_h(\mathbf{x})|\mathcal{D}, \tilde{\mathbf{z}}) 
   = &  \; \mathbb{E}(f_h(\mathbf{x})|\mathcal{D}) + 
   \Big( \frac{{\rm{cov}}(f_h(\mathbf{x}), f_h(\tilde{\mathbf{x}})| \mathcal{D})}{{\rm{var}}(f_h(\tilde{\mathbf{x}})|\mathcal{D})}
\nonumber \\
    &  * \big(\overline{f}_h(\tilde{\mathbf{x}})-\mathbb{E}(f_h(\tilde{\mathbf{x}})|\mathcal{D})\big)\Big)
\nonumber \\
    = & \;\mathbb{E}(f_h(\mathbf{x})|\mathcal{D}),
\label{recur_mean} \\
    {\rm{var}}\big(f_h(\mathbf{x})|\mathcal{D}, \tilde{\mathbf{z}})\big) = & \; {\rm{var}}(f_h(\mathbf{x})|\mathcal{D}) -
    \frac{{\rm{cov}}(f_h(\mathbf{x}),
    f_h(\tilde{\mathbf{x}})| \mathcal{D})^2}
    {{\rm{var}}(f_h(\tilde{\mathbf{x}})|\mathcal{D})}.
\label{recur_cov}
\end{align} \end{linenomath} 
\normalsize

Up to this point, the algorithm for single-fidelity method can be considered complete, and one simply needs to find $\tilde{\mathbf{x}}$ to minimize \eqref{ivr}. However, for the purpose of convenience in developing the bi-fidelity method later, it is more desirable to formulate an equivalent acquisition through the reduction of the variance, i.e., the benefits, of adding a hypothetical sample. This can be expressed as
\begin{equation}
    B(\tilde{\mathbf{x}})  = U(\mathcal{D}) - U(\mathcal{D}, \tilde{\mathbf{z}}),
\label{sgp_acq}
\end{equation}
where $U(\mathcal{D})$ is defined as \eqref{ivr} conditioning on $\mathcal{D}$ only. The next-best sample can then be selected through the solution of an optimization problem.
\begin{equation}
    \mathbf{x}^* = {\rm{argmax}}_{\tilde{\mathbf{x}} \in  \mathbb{R}^{d}} \; B(\tilde{\mathbf{x}}),
\label{opt_sgp}
\end{equation}
which can be directly solved using standard global optimization methods, e.g.,  multiple-starting L-BFGS-B quasi-Newton method \cite{nocedal1980updating} used in our study. {\color{black} The optimization \eqref{opt_sgp} is repeated for each sequential sample until reaching a user-defined number of samples $n_{lim}$ which needs to be practically chosen balancing the computational budget and required accuracy of the result.}

We finally summarize the full algorithm in Algorithm \textbf{1}.

\begin{algorithm}
    \caption{Single-fidelity method for CAV safety analysis}
  \begin{algorithmic}
    \REQUIRE Number of initial samples $n_{init}$, limit of number of samples $n_{lim}$
    \INPUT Initial dataset $\mathcal{D}=\{\mathbf{X}, \mathbf{Y} \}$
    \STATE \textbf{Initialization} 
        $n_{total} = n_{init}$
    \WHILE{$n_{total} < n_{lim}$}
      \STATE 1. Train the surrogate model with $\mathcal{D}$ to obtain \eqref{sgp1}
      \STATE 2. Solve the optimization \eqref{opt_sgp} to find the next-best sample $ \boldsymbol{x}^*$
      \STATE 3. Evaluate the function $f_h$ to get $f_h(\mathbf{x}^*)$
      \STATE 4. Update the dataset $\mathcal{D}$ with $\mathbf{X} = \mathbf{X} \; \cup \; \{\boldsymbol{x}^*\}$ and $ \mathbf{Y} = \mathbf{Y} \; \cup \; f_h(\mathbf{x}^*) $
      \STATE 5. $n_{total} = n_{total} + 1 $
    \ENDWHILE
\OUTPUT Compute $P_a$ according to \eqref{FP} based on the surrogate model \eqref{sgp2} 
  \end{algorithmic}
\label{al}
\end{algorithm}

\subsection{Bi-fidelity method}
We consider the situation that, in addition to the high-fidelity model $f_h$, we also have a low-fidelity model $f_l$ with lower computational cost. The model $f_l$ can be considered to provide an approximation to $f_h$ with a relation
\begin{equation}
    f_h(\mathbf{x}) = f_{l}(\mathbf{x}) + d(\mathbf{x}),
\label{AR}
\end{equation}
where $d(\mathbf{x})$ is an unknown difference function to be determined.  

We further assume that the cost for an evaluation using $f_h$ is $c_h$, and that for $f_l$ is $c_l$, with $c_h/c_l>1$. Since $f_l$ is associated with finite cost, we cannot assume that the full low-fidelity map is available to us, in contrast to the situation in  \cite{feng2020testing1, feng2021intelligent}. The adaptive sampling algorithm for this bi-fidelity application is required to find a sequence of samples with optimal fidelity level and location, i.e., $f_i(\mathbf{x})$ with $i=h$ or $l$ and $\mathbf{x}$ varying for each sample. For this purpose, the algorithm for the single-fidelity method needs to be extended in two aspects: (i) construction of the surrogate model through a bi-fidelity Gaussian process; and (ii) a more comprehensive acquisitive function measuring the benefit per computational cost for each sample, allowing the next-best sample to be selected in terms of both fidelity level and sampling position.

\subsubsection{surrogate model by BFGPR} Bi-fidelity Gaussian process regression (BFGPR) \cite{kennedy2000predicting} is a direct extension of GPR to infuse bi-fidelity data. Given a dataset $\mathcal{D}=\{\mathbf{X}, \mathbf{Y} \}$ consisting of two levels of model outputs $\mathbf{Y}= \{\mathbf{Y}_h, \mathbf{Y}_l\}$ at input positions $\mathbf{X}=\{\mathbf{X}_h, \mathbf{X}_l\}$, the purpose of the bi-fidelity Gaussian process is to learn the underlying relation $f_{h,l}(\mathbf{x})$ from $\mathcal{D}$. This can be achieved through an auto-regressive scheme, which models $f_h(\mathbf{x})$ in \eqref{AR} by two independent Gaussian processes $f_l(\mathbf{x}) \sim \mathcal{GP}(0, k_l(\mathbf{x},\mathbf{x}'))$ and $d(\mathbf{x}) \sim \mathcal{GP}(0, k_d(\mathbf{x},\mathbf{x}'))$. The posterior prediction $f_{h,l}(\mathbf{x})$ given the dataset $\mathcal{D}$ can then be derived as a Gaussian process:
\begin{equation}
    \begin{bmatrix} f_h(\mathbf{x}) \\ f_l(\mathbf{x}') \end{bmatrix} | \; \mathcal{D}  \sim \mathcal{N} \Big(\mathbb{E}\big(\begin{bmatrix} f_h(\mathbf{x}) \\ f_l(\mathbf{x}') \end{bmatrix}| \; \mathcal{D}\big), {\rm{cov}}\big(\begin{bmatrix} f_h(\mathbf{x}) \\ f_l(\mathbf{x}') \end{bmatrix}| \; \mathcal{D}\big) \Big),
\label{tfgp1}
\end{equation}
with mean and covariance respectively
\small
\begin{linenomath} \begin{align}
    \mathbb{E}(
    \begin{bmatrix} f_h(\mathbf{x}) \\ f_l(\mathbf{x}') \end{bmatrix}
   | \; \mathcal{D})  & = \rm{cov}(\begin{bmatrix} f_h(\mathbf{x}) \\ f_l(\mathbf{x}') \end{bmatrix}, \begin{bmatrix} \mathbf{Y}_h \\ \mathbf{Y}_l \end{bmatrix}) 
   {\rm{cov}}(\begin{bmatrix} \mathbf{Y}_h \\ \mathbf{Y}_l \end{bmatrix})^{-1} \begin{bmatrix} \mathbf{Y}_h \\ \mathbf{Y}_l \end{bmatrix},
\label{tfgp2} \\  
    {\rm{cov}}\big(\begin{bmatrix} f_h(\mathbf{x}) \\ f_l(\mathbf{x}') \end{bmatrix} | \; \mathcal{D} \big) & = {\rm{cov}}\big(\begin{bmatrix} f_h(\mathbf{x}) \\ f_l(\mathbf{x}') \end{bmatrix}) 
\nonumber \\    
    - \rm{cov}  ( \begin{bmatrix} f_h(\mathbf{x}) \\ f_l(\mathbf{x}') \end{bmatrix}, & \begin{bmatrix} \mathbf{Y}_h \\ \mathbf{Y}_l \end{bmatrix})  {\rm{cov}}(\begin{bmatrix} \mathbf{Y}_h \\ \mathbf{Y}_l \end{bmatrix})^{-1} \rm{cov}(\begin{bmatrix} \mathbf{Y}_h \\ \mathbf{Y}_l \end{bmatrix}, \begin{bmatrix} f_h(\mathbf{x}) \\ f_l(\mathbf{x}') \end{bmatrix}),
\label{tfgp3}
\end{align} \end{linenomath} 
\normalsize
where
\small
\begin{linenomath} \begin{align}
    & \rm{cov}( \begin{bmatrix} \mathbf{Y}_h \\ \mathbf{Y}_l \end{bmatrix})
    = 
    \begin{bmatrix}
      k_l(\mathbf{X}_h,\mathbf{X}_h) + k_d(\mathbf{X}_h,\mathbf{X}_h) &   k_l(\mathbf{X}_h,\mathbf{X}_l) \\
      k_l(\mathbf{X}_l,\mathbf{X}_h) &  k_l(\mathbf{X}_l,\mathbf{X}_l)
    \end{bmatrix},
\\
    & \rm{cov}( \begin{bmatrix} f_h(\mathbf{x}) \\ f_l(\mathbf{x}') \end{bmatrix}, \begin{bmatrix} \mathbf{Y}_h \\ \mathbf{Y}_l \end{bmatrix}) 
    = 
    \begin{bmatrix}   k_l(\mathbf{x}, \mathbf{X}_h) + k_d(\mathbf{x}, \mathbf{X}_h) &    k_l(\mathbf{x}, \mathbf{X}_l) \\    k_l(\mathbf{x}', \mathbf{X}_h)  &   k_1(\mathbf{x}', \mathbf{X}_l)
    \end{bmatrix}, 
\\
    & {\rm{cov}}\big(\begin{bmatrix} f_h(\mathbf{x}) \\ f_l(\mathbf{x}') \end{bmatrix})
    = 
    \begin{bmatrix}    k_l(\mathbf{x}, \mathbf{x}) + k_d(\mathbf{x}, \mathbf{x}) &    k_l(\mathbf{x}, \mathbf{x}') \\    k_l(\mathbf{x}', \mathbf{x})  & k_l(\mathbf{x}', \mathbf{x}')
    \end{bmatrix}. 
\end{align} \end{linenomath} 
\normalsize

We note that $f_h(\mathbf{x})|\mathcal{D}$ in \eqref{tfgp1}, as the major prediction in BFGPR, provides the high-fidelity function infusing both high and low fidelity samples $\{\mathbf{X}, \mathbf{Y} \}$ (instead of only $\{\mathbf{X}_h, \mathbf{Y}_h \}$). This is achieved, intuitively, through the two Gaussian processes on $f_l(\mathbf{x})$ and $d(\mathbf{x})$ which rely on all data. The prediction $f_h(\mathbf{x})|\mathcal{D}$ will be used as the surrogate model for the computation of $P_a$ and the development of acquisition function.  

\subsubsection{bi-fidelity acquisition function}

In the bi-fidelity context, the next-best sample needs to be determined in terms of both its location and fidelity level. Given a total cost budget, the principle to select the next-best sample is to maximize its benefit per cost. Accordingly, we consider the optimization of an acquisition function which captures both the benefit and cost of a sample $\tilde{\mathbf{x}}$:
\begin{equation}
    \mathbf{x}^*, i^* = {\rm{argmax}}_{\tilde{\mathbf{x}} \in  \mathbb{R}^{d}, i \in \{h, l\}} \; B_i(\tilde{\mathbf{x}}) / c_i.
\label{opt}
\end{equation}
Following the formulations in the single-fidelity problem, the benefit of adding an $i$-fidelity hypothetical sample at $\tilde{\mathbf{x}}$, $B_i(\tilde{\mathbf{x}})$, can be expressed as
\begin{equation}
    B_i(\tilde{\mathbf{x}})   = U(\mathcal{D}) - U(\mathcal{D}, \tilde{\mathbf{z}}_i), \quad i = h, l,
\label{tfgp_acq}
\end{equation}
with $\tilde{\mathbf{z}}_i = \{\tilde{\mathbf{x}}, \overline{f}_i(\tilde{\mathbf{x}})\}$ and 
\begin{equation}
    U(\mathcal{D}, \tilde{\mathbf{z}}_i) = \int \mathrm{var}^{\frac{1}{2}} \Big(\mathbf{1}_{\delta}\big(f_h(\mathbf{x})|\mathcal{D}, \tilde{\mathbf{z}}_i\big)\Big) p_{\mathbf{x}}(\mathbf{x}){\rm{d}}\mathbf{x}.
\label{ivr_bf}
\end{equation}

The computation of \eqref{ivr_bf} can be conducted following \eqref{recur_mean} and \eqref{recur_cov} adapted to the bi-fidelity context using the BFGPR surrogate model. In solving \eqref{opt} as a combined discrete and continuous optimization problem, we first find the optimal location $\mathbf{x}$ for each fidelity $i$, i.e., $\mathbf{x}_i^*= {\rm{argmax}}_{\tilde{\mathbf{x}} \in  \mathbb{R}^{d}} \; B_i(\tilde{\mathbf{x}})$ for $i=h,l$, then we compare the benefit per cost $B_i(\mathbf{x}_i^*)/c_i$ between $i=h$ and $i=l$ and find the optimal fidelity level $i^*$, i.e., $i^*= {\rm{argmax}}_{i \in \{h,l\}} \; B_i(\mathbf{x}_i^*) / c_i$. We further remark that this idea of maximizing benefit per cost has been systematically tested for another sampling purpose, i.e., with a different benefit function $B_i(\tilde{\mathbf{x}})$, in \cite{gong2022multi}.

We finally summarize the full algorithm in Algorithm \textbf{2}.

\begin{algorithm}
    \caption{Bi-fidelity method for CAV safety analysis}
  \begin{algorithmic}
    \REQUIRE Number of initial samples $\{n^{init}_h, n^{init}_l\}$, cost of each fidelity model $\{c_h, c_l\}$, total cost budget $c_{lim}$
    \INPUT Initial dataset $\mathcal{D}=\{\mathbf{X}, \mathbf{Y} \}$ with $\mathbf{X}=\{\mathbf{X}_h, \mathbf{X}_l\}$ and $\mathbf{Y}=\{\mathbf{Y}_h, \mathbf{X}_l\}$
    \STATE \textbf{Initialization} 
        $c_{total} = n^{init}_h \; c_{h} + n^{init}_l \; c_{l}$
    \WHILE{$c_{total} < c_{lim}$}
      \STATE 1. Train the surrogate model with $\mathcal{D}$ to obtain \eqref{tfgp1}
      \STATE 2. Solve the optimization \eqref{opt} to find the next-best sample $\{i^*, \boldsymbol{x}^*\}$
      \STATE 3. Evaluate the $i^*-$fidelity function to get $f_{i^*}(\mathbf{x}^*)$
      \STATE 4. Update the dataset $\mathcal{D}$ with $\mathbf{X}_{i^*} = \mathbf{X}_{i^*} \cup  \{\boldsymbol{x}^*\}$ and $ \mathbf{Y}_{i^*} = \mathbf{Y}_{i^*} \cup \{f_{i^*}(\mathbf{x}^*)\}$
      \STATE 5. $c_{total} = c_{total} + c_{i^*} $
    \ENDWHILE
\OUTPUT Compute the $P_a$ according to \eqref{FP} based on the surrogate model \eqref{tfgp2} 
  \end{algorithmic}
\label{al}
\end{algorithm}

\section{Cut-in case analysis}
In this section, we demonstrate the application of our proposed (single and bi-fidelity) methods to the cut-in problem, starting with a more detailed description of the setup of the case. Since our method is new to the general reliability analysis field, we also document its favorable performance for two widely-used benchmark problems in reliability analysis in Appendix C.

\subsection{Case setup}  \label{case}

\begin{figure}[htbp]
     \centering 
     \begin{minipage}[b]{\linewidth}   
         \includegraphics[width=7.5cm]{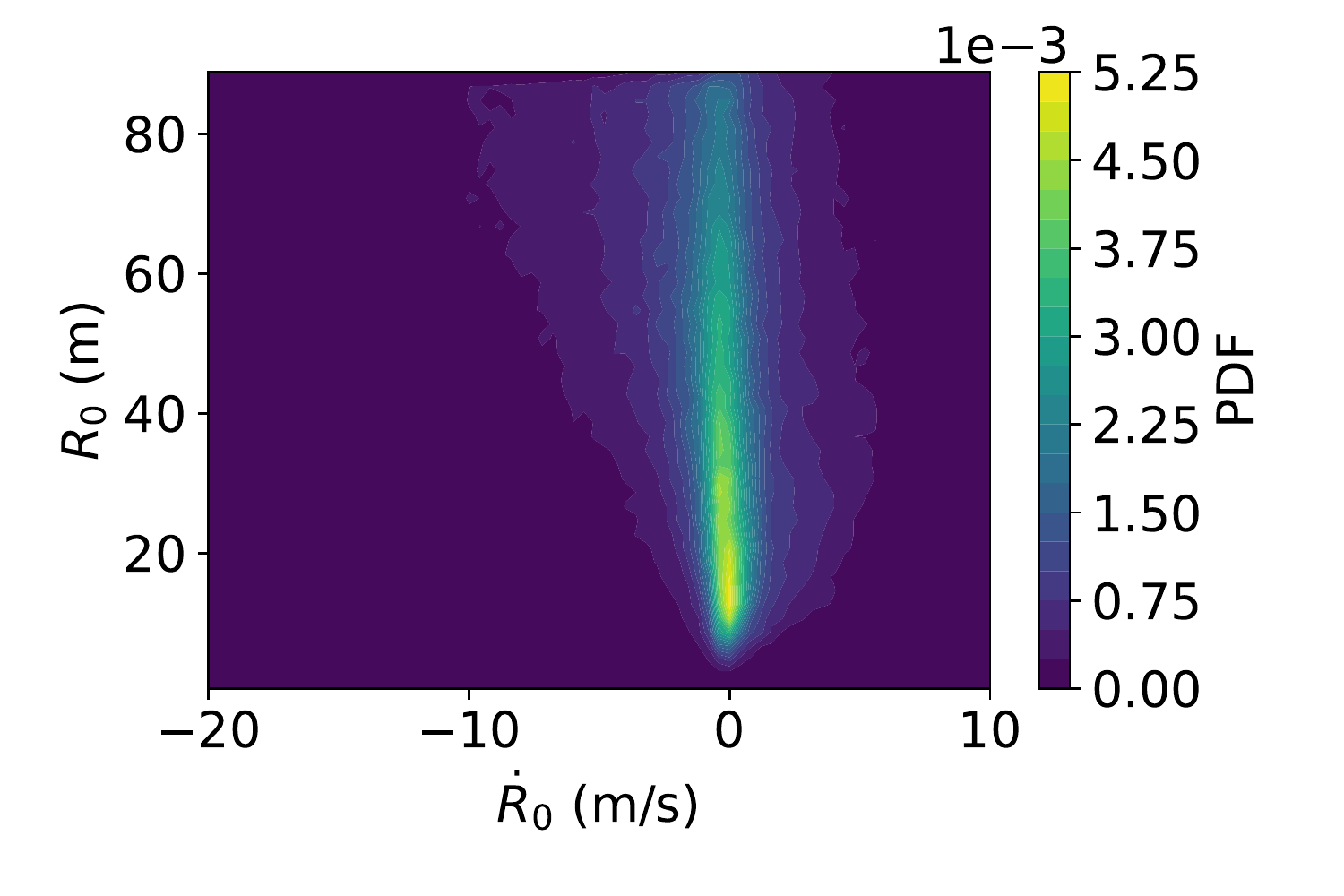} 
         \centering{\\ (a)}
     \vspace{0.2cm}
     \end{minipage}
     \begin{minipage}[b]{\linewidth}         
         \includegraphics[width=7cm]{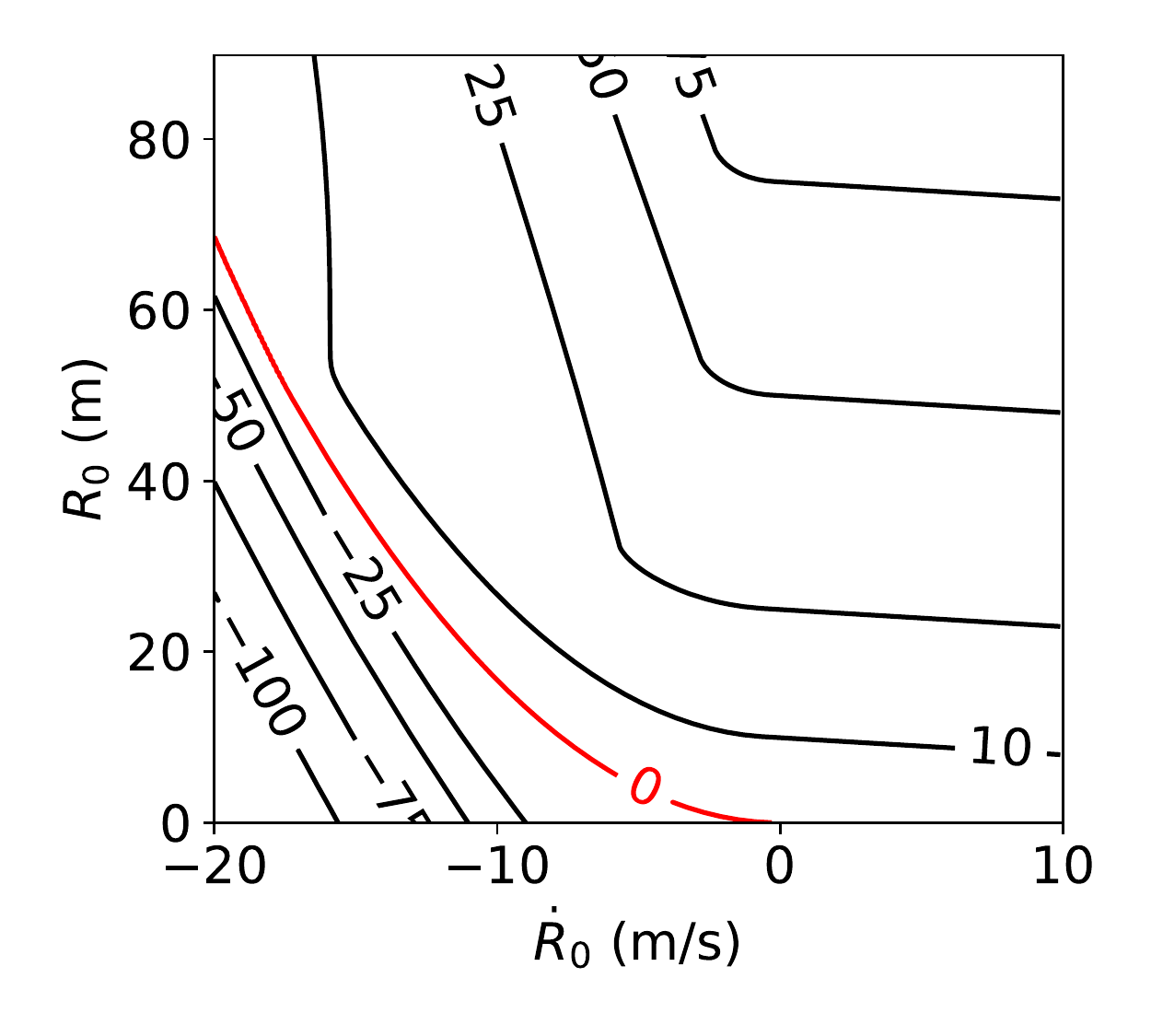}
         \centering{\\(b)}
     \vspace{0.2cm}
     \end{minipage}
        \caption{(a) probability distribution and (b) output from $f_h$ for input parameters $R_0$ and $\dot{R}_0$. The limiting state $\{\mathbf{x}: f_h(\mathbf{x}) = 0\}$ is marked in (b) by a red line.}
\label{fig:cutin_true}
\end{figure}

The cut-in situation is illustrated in Fig. \ref{fig:cut-in} where a BV makes a line change in front of a CAV. We assume that the BV moves in a constant speed $u_{BV}=20m/s$ after the cut-in moment, so that (given the CAV model) the performance of the CAV only depends on $\mathbf{x} = (R_0, \dot{R}_0)$, the initial range and range rate at the cut-in moment $t=0$ (time  $t$ in unit of seconds hereafter). The probability of $\mathbf{x}$ is generated from the naturalistic driving data (NDD) of the Safety Pilot Model Deployment (SPMD) at the University of Michigan \cite{bezzina2014safety}. A total
number of 414,770 qualified cut-in events are analyzed with joint distribution of $\mathbf{x}$ shown in Fig. \ref{fig:cutin_true} (a).

The model output for the problem is the minimum range between two vehicles during their speed adjustment process for $t\geq 0$. In this work, we use the the Intelligent Driving Model (IDM) which describes the speed of CAV by an ordinary differential equation
\begin{equation}
    \frac{d u_{cav}(t)}{d t} =  \alpha \Big( 1 - \big( \frac{u_{cav}(t)}{\beta} \big)^{c} - \big( \frac{s(u_{cav}(t), \dot{R}(t))}{R(t) - L} \big)^2
    \Big),
\label{IDM_model_1} \\
\end{equation}
\begin{equation}
     s(u_{cav}(t), \dot{R}(t)) =  s_0  + u_{cav}(t)  T + \frac{u_{cav}(t) \dot{R}(t)}{2 \sqrt{\alpha b}},
\label{IDM_model_2}
\end{equation} 
where $\alpha$, $\beta$, $c$, $s_0$, $L$, $b$, and $T$ are constant parameters and values in \cite{ro2017formal} are used here\footnote{$\alpha=2$, $\beta=18$, $c=4$, $s_0=2$, $L=4$, $b=3$, and $T=1$.}. We integrate \eqref{IDM_model_1} in time using forward Euler method starting from initial condition $u_{cav}(t=0)=u_{BV} - \dot{R}_0$, and accordingly find the range $R$ and range rate $\dot{R}$ for $t\geq 0$. In addition, we constrain the velocity and acceleration of the CAV to be $2 \leq u_{cav}(t) \leq 40 \, m/s$ and $-4 \leq d u_{cav}(t) / d t \leq 2  \, m/s^2$. The minimum range $R$ obtained for $0\leq t\leq 10$ is taken as the model output. We use a number of different time resolutions in integrating \eqref{IDM_model_1}, with the result obtained for $\Delta_t=0.2$ as the high-fidelity model output ($f_h$) that is plotted in Fig. \ref{fig:cutin_true} (b) for visualization. The results for $\Delta_t = 0.5, 1, 2,$ and 5 are taken as the outputs from low-fidelity models ($f_l$) with different fidelity levels, i.e., we will consider the bi-fidelity context as $f_h$ plus one of the low-fidelity options $f_l$. The cost ratio $c_h/c_l$ is inversely proportional to the ratio of $\Delta_t$'s. 

In order to evaluate the performance of our methods in Sec. \ref{results}, we compute accurate (ground truth) values of $P_a$ for both $\delta=0$ and $\delta=3$ (see \eqref{indicator}) according to high-fidelity model $f_h$ for all 414,770 available events. The former will be used for validating the single-fidelity method (so that its performance can be compared to existing cases in literature) and the latter for the bi-fidelity method (to have a considerable difference between the results from $f_h$ and $f_l$).

\subsection{Results} \label{results}
\subsubsection{single-fidelity results} \label{single}
\begin{figure}
    \centering
    \includegraphics[width=7.5cm]{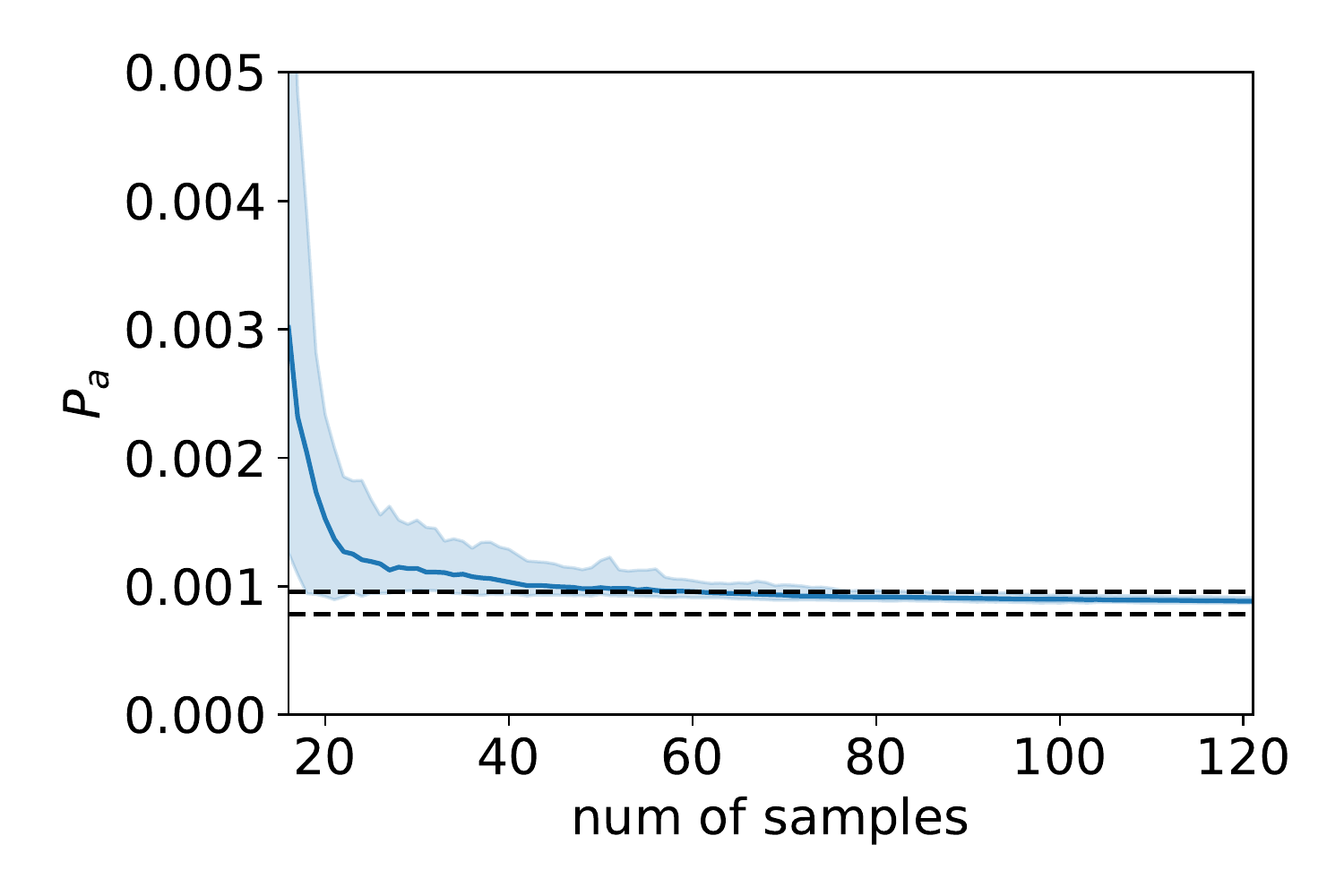}
    \caption{Results of $P_a$ from the single-fidelity method, presented by the median value (\blueline) as well as the 15\% and 85\% percentiles (shaded region) from 200 experiments. The true solution of $P_a$ (\blackdashedline) is shown in terms of the 10\% error bounds.}
    \label{fig:cutin_res_1}
\end{figure}

We first apply the single-fidelity method (Algorithm \textbf{1}) to the context where only $f_h$ (IDM with $\Delta_t = 0.2$) is available to us. The computation starts from 16 initial random samples, with the following 104 adaptive samples obtained from Algorithm \textbf{1}. The results of $P_a$ (estimated from \eqref{FP} in the sampling process) as a function of the number of samples are shown in Fig. \ref{fig:cutin_res_1} together with the 10\% error bounds of the ground truth. Since the value of $P_a$ depends on the locations of the initial samples, we quantify its uncertainty by plotting both the median value as well as 15\% and 85\% percentiles in Fig. \ref{fig:cutin_res_1} obtained from 200 applications of our algorithm starting from different initial samples. The percentile concept is used here because the distribution of $P_a$ over different experiments is not guaranteed to be Gaussian, and the 15\% and 85\% percentiles are used for the convenience of a fair comparison with other results discussed below.

From Fig. \ref{fig:cutin_res_1} we see that it only takes 83 samples (or 67 adaptive samples) for the upper and lower percentiles to converge into the 10\% error bounds of the ground truth. In comparison, to reach convergence with a similar criterion\footnote{Our convergence criterion means that within every 100 experiments, 70 of them provide results within 10\% error bounds. This is, by definition, equivalent to the relative half-width of the 70\% confidence interval falling below 0.1. The latter, as described in Fig. 9 of \cite{feng2020testing3} (presented in terms of $\pm 2$ standard deviation which is close to two times of the 70\% confidence interval for Gaussian estimation in importance sampling), takes 121 samples.}, it takes 121 samples for the importance sampling method presented in earlier work \cite{feng2020testing3}. It should also be emphasized that the method in \cite{feng2020testing3} requires a pre-known low-fidelity map (e.g., $f_l$ with negligible cost) to guide the proposal distribution in importance sampling. This extra component is not required at all in our method. Therefore, for this validation case, it can be concluded that our approach takes about 2/3 number of samples to achieve the same accuracy as \cite{feng2020testing3} based on the current criterion, and can be conducted in a much simpler setting. Another possible source of comparison is \cite{sun2021adaptive}, which performs the adaptive sampling with a simplified/empirical acquisition function, without needing a low-fidelity model, to a slightly more complicated 3D car-following problem. It takes O(500) samples to obtain convergent result of accident rate, which is based on one experiment result without analyzing the uncertainty bounds as we do here. It is desirable to apply our method to the same case, but the information provided in \cite{sun2021adaptive} is not sufficient for us to do so (e.g., no code or input probability data is provided). {\color{black} We further emphasize that all methods reviewed here feature different computational complexity in selecting the samples. However, the total cost in real applications is generally dominated by the cost to evaluate the car performance for each sample, which can come from expensive simulations or on-road/closed-facility tests. Therefore, the total cost is simply controlled by the number of samples to reach certain level of accuracy in the accident rate, as we discussed above. }

\begin{figure}
    \centering
    \includegraphics[width=7cm]{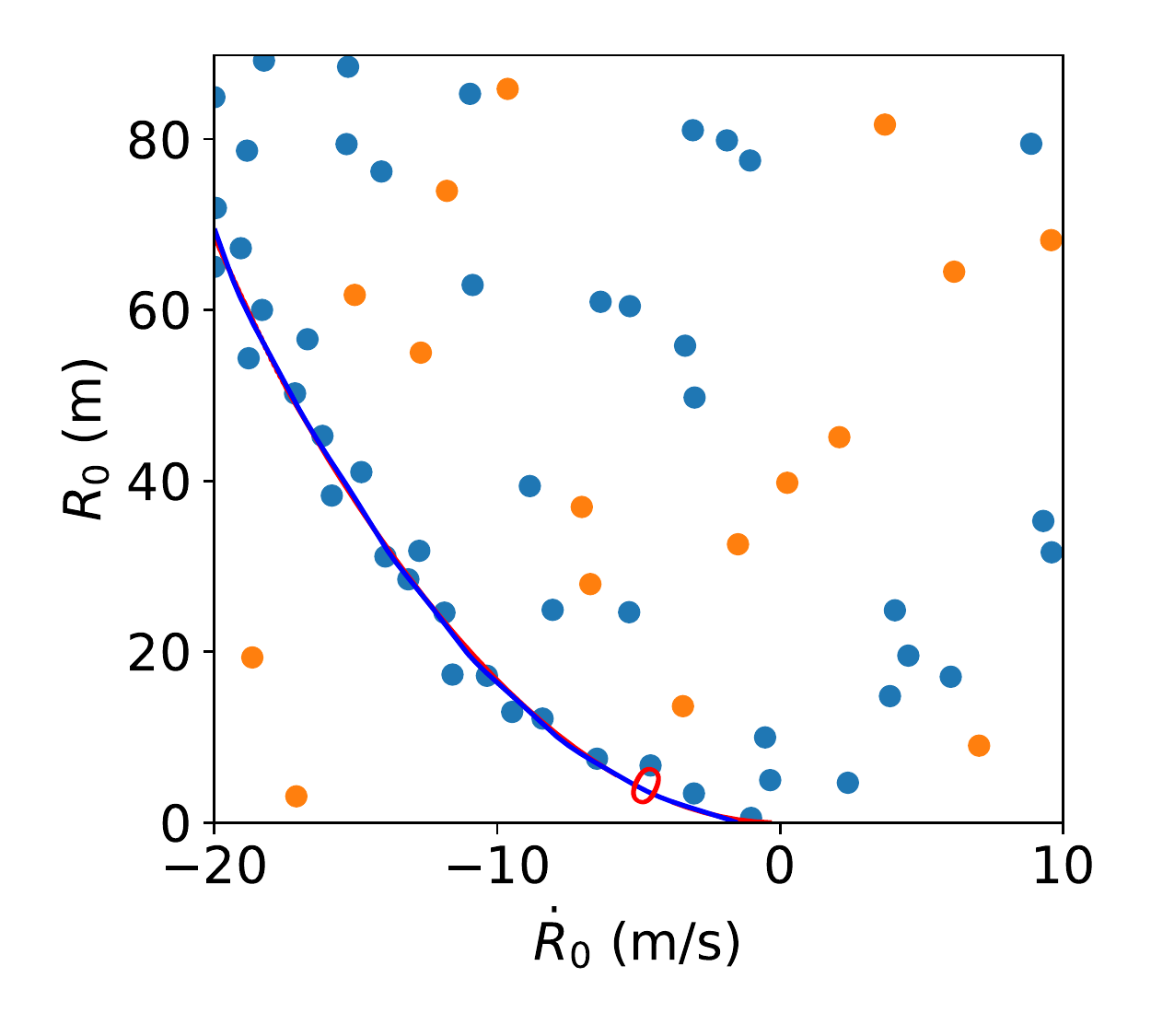}
    \caption{Positions of 16 initial samples (\tikzcircle{2pt, Orange}) and 50 adaptive samples (\tikzcircle{2pt,NavyBlue}) from a typical experiment of our method, as well as the learned limiting state (\blueline) compared to the exact one (\redline).}
    \label{fig:cutin_sampling_1}
\end{figure}

Finally, we plot in Fig. \ref{fig:cutin_sampling_1} the sampling positions in the input space $\mathbf{x}$ for a typical case out of the 200 experiments. After 16 initial random samples, we see that most adaptive samples are located close to the limiting state to better resolve $P_a$. The limiting state $\{\mathbf{x}: f_h(\mathbf{x}) = 0\}$ estimated from the GPR constructed by only 50 adaptive samples is also included in the figure to show its proximity to the true state. 

\begin{figure*}
\centering
\begin{minipage}[b]{0.375\linewidth}
\includegraphics[width = \linewidth]{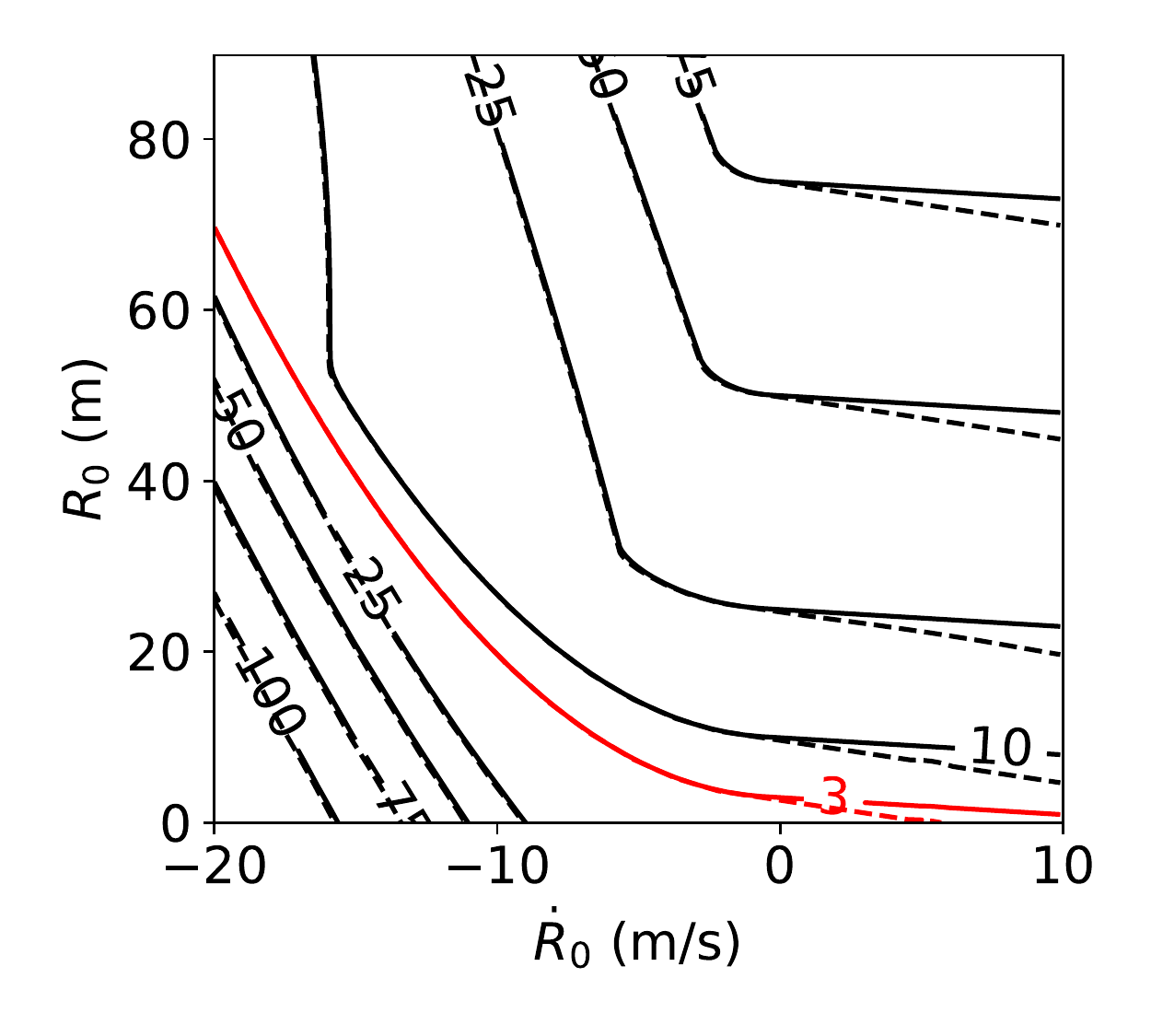}
\centering{\quad (a)}
\end{minipage}
\begin{minipage}[b]{0.375\linewidth}
\includegraphics[width = \linewidth]{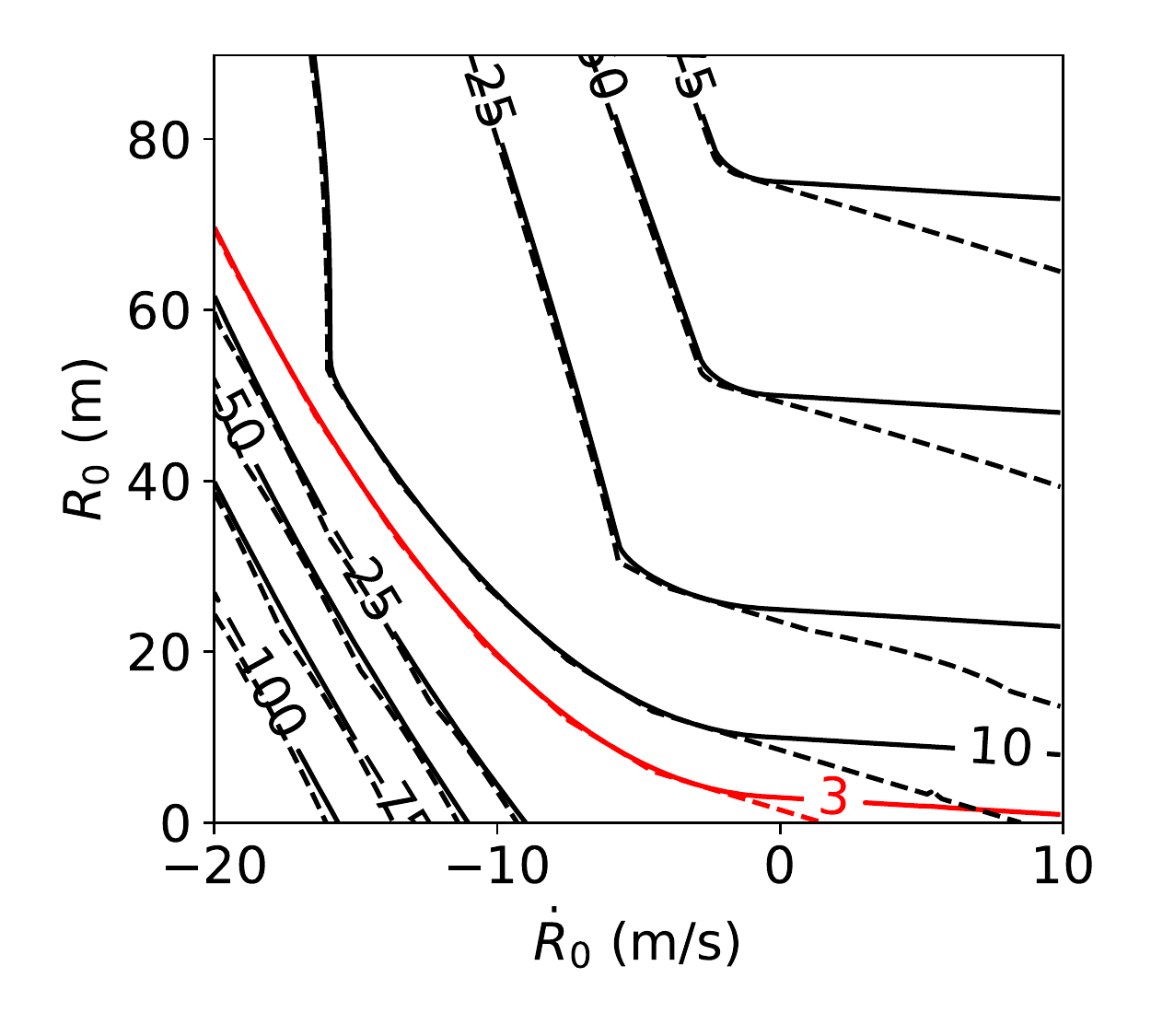}
\centering{\quad (b)}
\end{minipage}
\begin{minipage}[b]{0.375\linewidth}
\includegraphics[width = \linewidth]{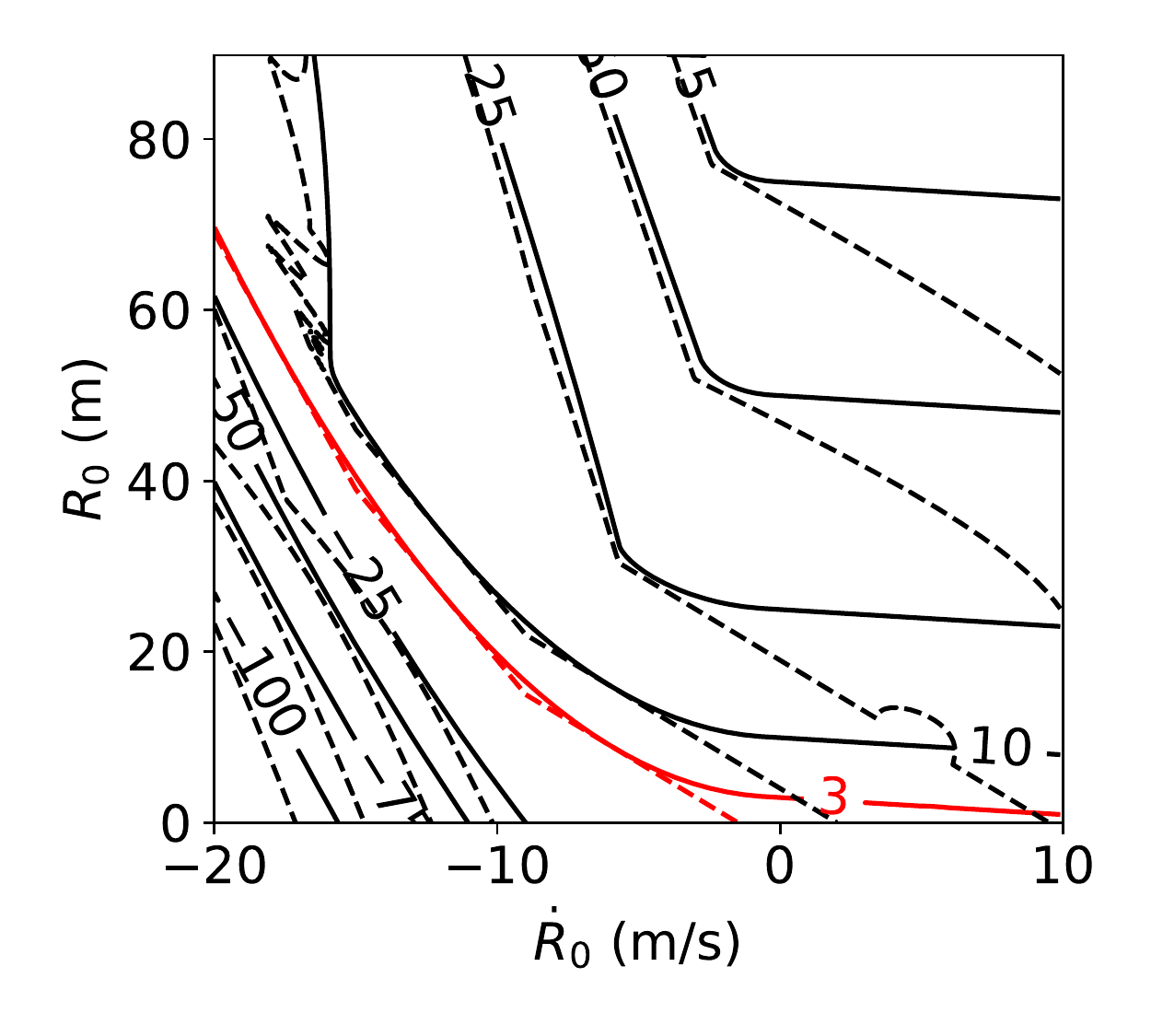}
\centering{\quad (c)}
\end{minipage}
\begin{minipage}[b]{0.375\linewidth}
\includegraphics[width = \linewidth]{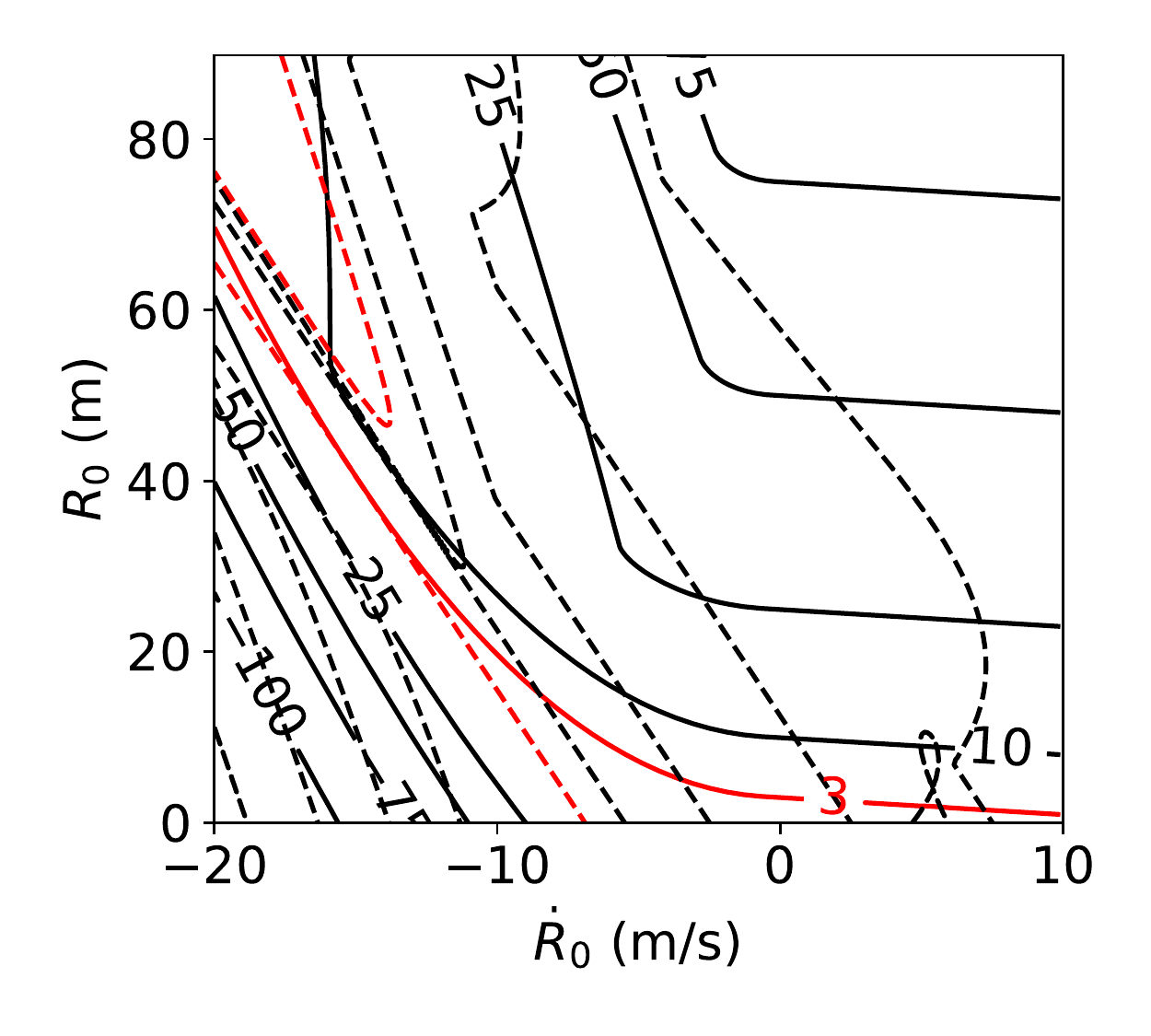}
\centering{\quad (d)}
\end{minipage}
\caption{Output contour from $f_l$ (\blackdashedline) of (a) $\Delta_t=0.5$, (b) $\Delta_t=1$, (c) $\Delta_t=2$, (d) $\Delta_t=5$ compared with contour from $f_h$ (\blackline) of $\Delta_t=0.2$ . The limiting states $\{\mathbf{x}: f_{h,l}(\mathbf{x}) = 3\}$ of $f_h$ and $f_l$ are respectively marked by (\redline) and (\reddashedline).}
\label{idm_2}
\end{figure*}

\begin{figure*}
\centering
\begin{minipage}[b]{0.4\linewidth}
\includegraphics[width = \linewidth]{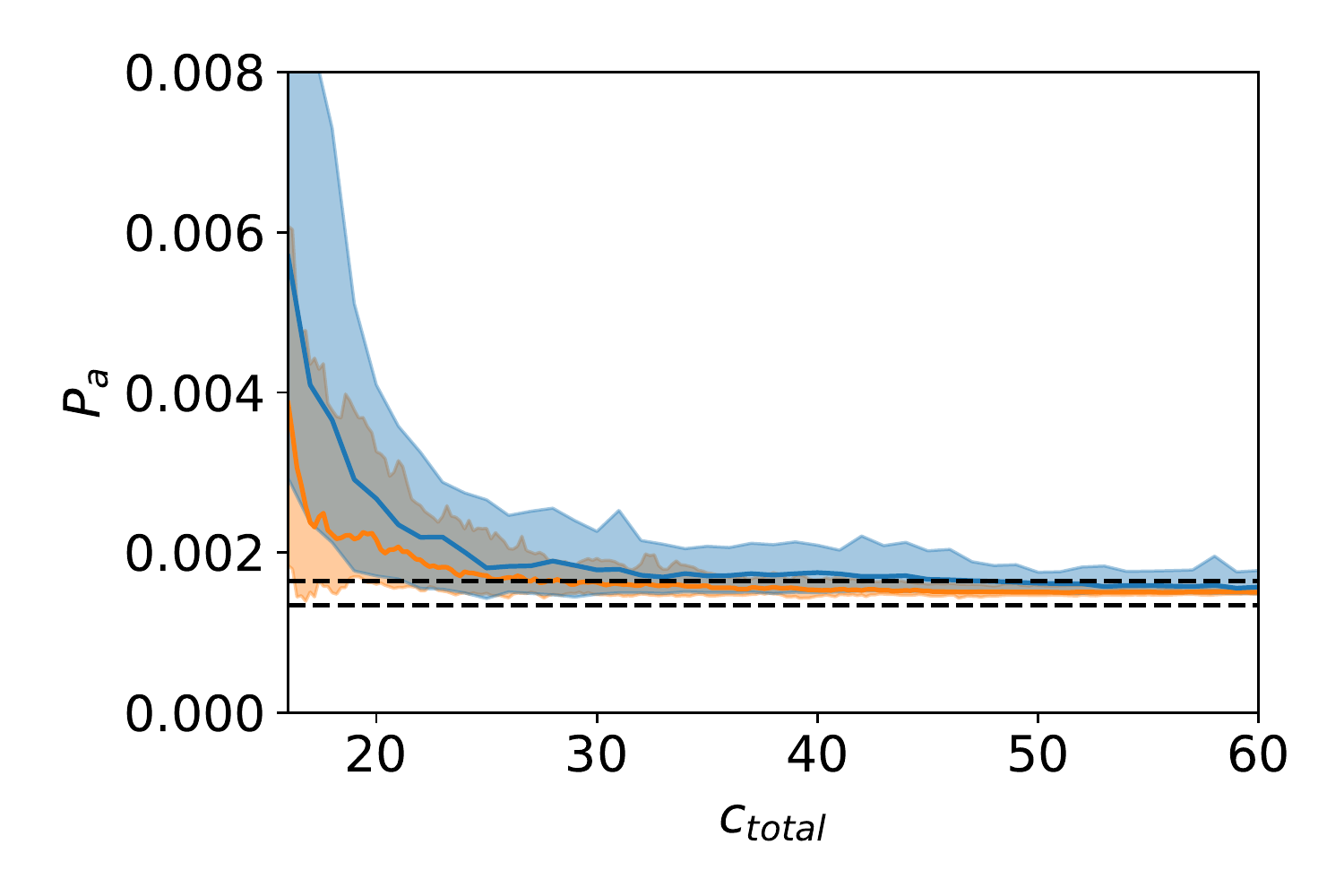}
\centering{\quad (a)}
\end{minipage}
\begin{minipage}[b]{0.4\linewidth}
\includegraphics[width = \linewidth]{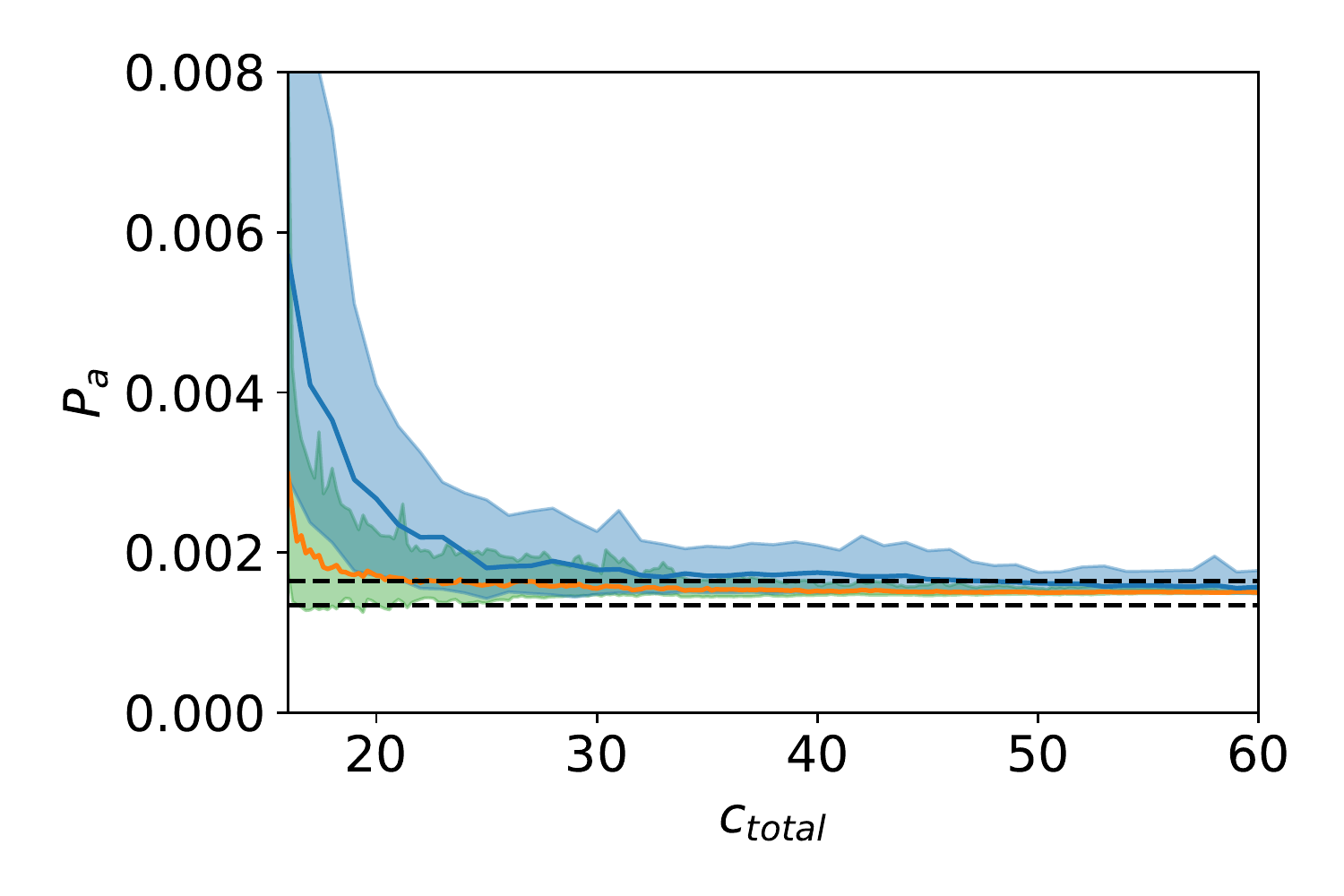}
\centering{\quad (b)}
\end{minipage}
\begin{minipage}[b]{0.4\linewidth}
\includegraphics[width = \linewidth]{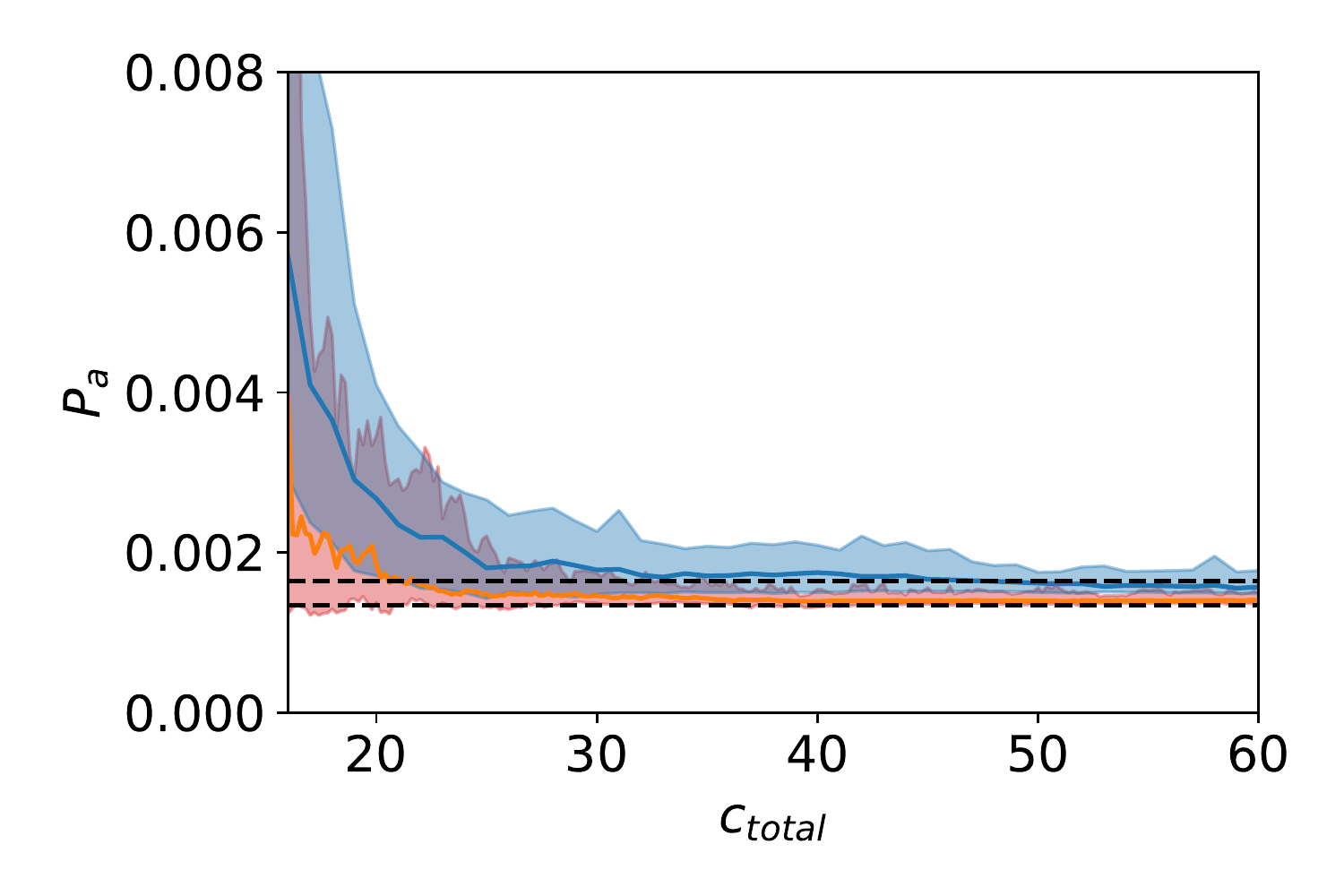}
\centering{\quad (c)}
\end{minipage}
\begin{minipage}[b]{0.4\linewidth}
\includegraphics[width = \linewidth]{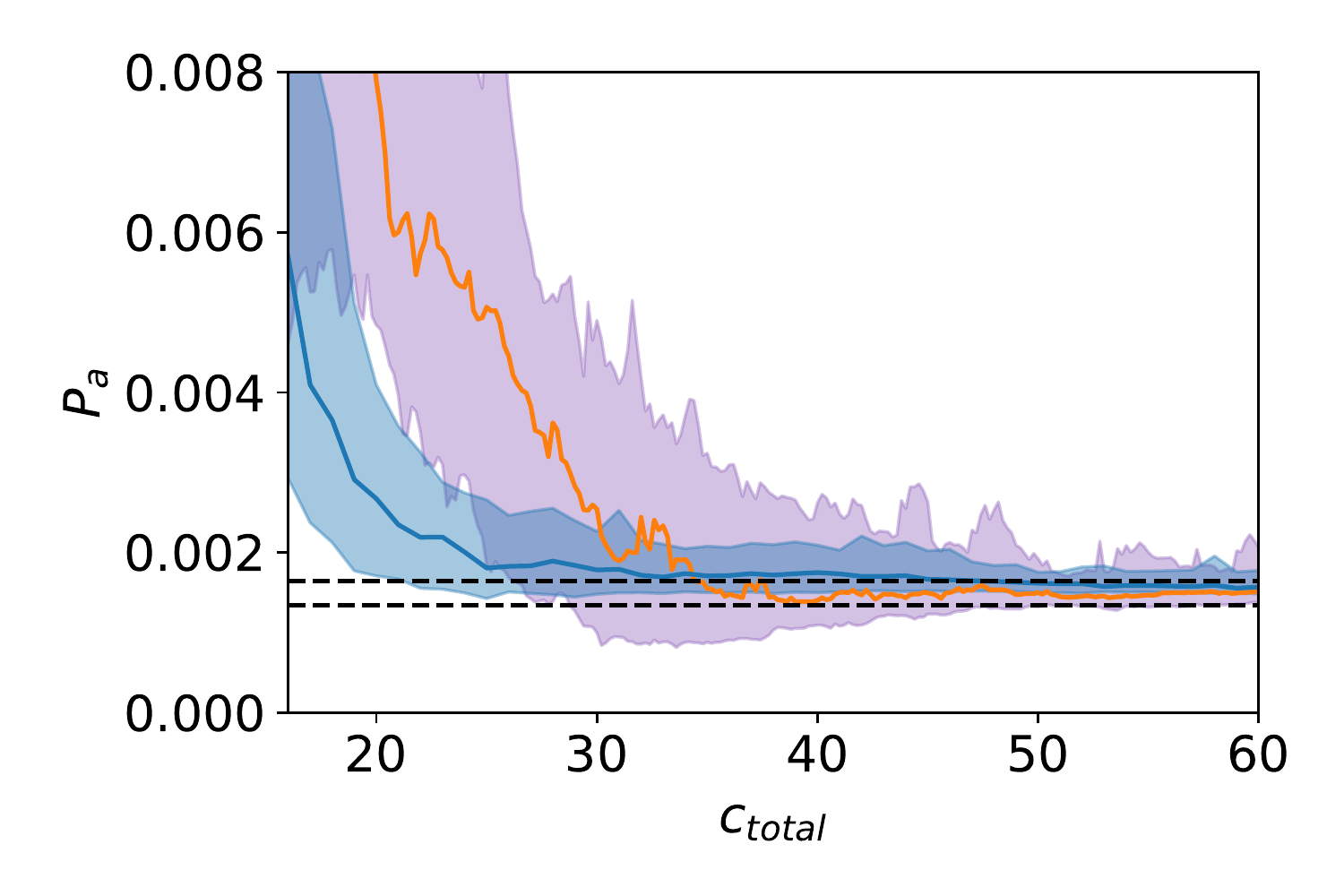}
\centering{\quad (d)}
\end{minipage}
 \caption{Results of $P_a$ from the single-fidelity method with $f_h$ (\blueline) and bi-fidelity method with both $f_h$ and $f_l$ ($\Delta_t=0.5$: (\orangeline) in (a); $\Delta_t=1$: (\greenline)  in (b), $\Delta_t=2$: (\redline)  in (c);, $\Delta_t=5$: (\purpleline)  in (d)), presented by the median value (solid lines) as well as the 15\% and 85\% percentiles (shaded region) from 50 experiments. The true solution of $P_{f}$ (\blackdashedline) is shown in terms of the 10\% error bounds.}
\label{fig:cutin_res_3}
\end{figure*}

\begin{table}[h]
\begin{center}
\caption{Accident rates computed by models with different time resolutions}
\begin{threeparttable}[t]
\begin{tabular}{|c|c|c|c|c|c|}
\hline
$\Delta_t$ & 0.2                                                        & 0.5                                                       & 1                                                         & 2                                                         & 5                                                          \\ \hline
$P_a$      & \begin{tabular}[c]{@{}c@{}}0.001489\\ \;  \end{tabular} & \begin{tabular}[c]{@{}c@{}}0.001375\\ (8\%)\tnote{1} \end{tabular} & \begin{tabular}[c]{@{}c@{}}0.001202\\ (19\%)\end{tabular} & \begin{tabular}[c]{@{}c@{}}0.001114\\ (25\%)\end{tabular} & \begin{tabular}[c]{@{}c@{}}0.004725\\ (217\%)\end{tabular} \\ \hline
\end{tabular}
\begin{tablenotes}
     \item[1] Relative differences are computed with respect to the result of $\Delta_t=0.2$.
\end{tablenotes}
\end{threeparttable}
\label{tab:P_a}
\end{center}
\end{table}

\begin{table}[h]
\caption{Summary of the performance of single-fidelity (No. 1) and bi-fidelity (No. 2-5) cases}
\begin{threeparttable}[t]
\begin{tabular}{|c|c|c|c|c|c|c|}
\hline
No.    &   $\Delta_t(f_h)$   & $\Delta_t(f_l)$ & $c_h/c_l$   & $n_l/n_h$ \tnote{1} & percentiles \tnote{2} & median \tnote{2} \\ \hline
1            &  0.2    & $-$        & $-$        & - & 82  & 47 \\ \hline
2            &  0.2    & 0.5        & 2.5        & 6.63 & 42   & 30 \\ \cline{1-7} 
3            &  0.2    & 1          & 5          & 6.22 & 38   & 22\\  \cline{1-7} 
4            &  0.2    & 2          & 10         & 8.15 & 39   &  23\\ \cline{1-7}
5            &  0.2    & 5          & 25         & 4.1 & $>60$ & 35 \\ \hline
\end{tabular}
\begin{tablenotes}
     \item[1] $n_l/n_h$ gives the ratio of low and high-fidelity sampling numbers in the adaptive sampling process.
     \item[2] ``percentiles'' and ``median'' columns respectively give the units of cost for the (15\% and 85\%) percentiles and median of $P_a$ to converge to the 10\% error bounds of the ground truth.
\end{tablenotes}
\end{threeparttable}
\label{tab:summary}
\end{table}

\subsubsection{Bi-fidelity results} We next consider the bi-fidelity application where the high-fidelity model $f_h$ is used together with a low-fidelity model $f_l$. Different choices of $f_l$ are considered, which are obtained from IDM model with coarser time resolutions of $\Delta_t=0.5, 1, 2$, and 5, in contrast to $\Delta_t=0.2$ for $f_h$. As shown in Fig. \ref{idm_2}, these low-fidelity models provide different levels of approximation to $f_h$, especially regarding the limiting state $\{\mathbf{x}: f_h(\mathbf{x}) = 3\}$ considered here. While the $f_l$ of $\Delta_t=0.5$ provides a close estimation of the limiting state, the $f_l$ of $\Delta_t=5$ provides a poor estimation even including an extra region that leads to false accidents. The values of $P_a$ computed solely by different choices of $f_l$ are listed in Table \ref{tab:P_a}, which shows $8-217\%$ relative difference with the ground-truth value (computed by $f_h$) for $f_l$ with $\Delta_t=0.5-5$. {\color{black} In particular, results obtained with $\Delta_t=5$ show an abnormally large $P_a$ because of the false-accident region in Fig. \ref{idm_2}(d).}  In terms of computational cost, we consider that one application of $f_h$ takes one unit of cost, and one application of $f_l$ takes $0.2/\Delta_t$ units of cost.

The results of our bi-fidelity method with respect to total cost $c_{total}$ in the sampling process are shown in Fig. \ref{fig:cutin_res_3} for four cases, with (a)-(d) corresponding to $f_l$ of $\Delta_t = 0.5, 1, 2,$ and 5. In each bi-fidelity case, the application starts with 8 high-fidelity samples (8 units of cost) and 8 $c_h/c_l$ low-fidelity samples (8 units of cost) as the initial dataset, followed by adaptive samples with 44 units of cost. Also shown in Fig. \ref{fig:cutin_res_3} are single-fidelity results using $f_h$, starting from 16 high-fidelity random samples (i.e., 16 units of cost), as well as 10\% error bounds of the ground-truth $P_a$. Similar to Sec. \ref{single}, the results are presented including the median value as well as 15\% and 85\% percentiles obtained from 50 applications. We see that in general, the bi-fidelity approach converges much faster than the single-fidelity approach, especially regarding the cases with $f_l$ constructed by $\Delta_t=0.5, 1,$ and 2 in (a), (b) and (c). For these cases, the percentiles of $P_a$ converge into the 10\% error bound in O(40) units of cost, whereas the single-fidelity approach takes more than 60 units (82 units upon extended test not shown in the Fig.) to reach the same convergence criterion. For the bi-fidelity case with $f_l$ of $\Delta_t=5$, the percentiles of $P_a$ converge with a comparable (but slightly slower) speed relative to the single-fidelity case within 60 units of cost as seen in (d). However, even in this case, the median of $P_a$ in the bi-fidelity approach converges to the 10\% error bound in O(35) samples that is faster than O(50) samples in the single-fidelity approach. It must also be kept in mind that for this bi-fidelity case, $P_a$ starts with a much less accurate value at the beginning of the adaptive sampling process, due to the initial $8 c_h / c_l = 200$ low-fidelity samples that are very misleading in constructing the initial surrogate model. Shall a different allocation of high/low fidelity samples are used for the initial dataset, it is possible to further improve the performance of this bi-fidelity application.

We further summarize the performance of all 5 cases (one single-fidelity and four bi-fidelity cases) in Table \ref{tab:summary}. In addition to the exact units of cost consumed for convergence, a notable information is provided by the column of $n_l/n_h$ that gives the ratio of low and high-fidelity sampling numbers in the adaptive sampling process. It is clear that $n_l/n_h$ does not monotonically increase with the increase of $c_h/c_l$, i.e., the algorithm does not select more low-fidelity samples just because they are cheaper but instead considers the benefit per cost of each sample formulated in \eqref{opt}. This is most evident when $c_h/c_l$ increases from 10 to 25 and meanwhile, $n_l/n_h$ drops from 8.15 to 4.1, mainly because a sample by $f_l$ with $\Delta_t=5$ does not provide much useful information to the computation of $P_a$. 

\begin{figure}
    \centering
    \includegraphics[width=7cm]{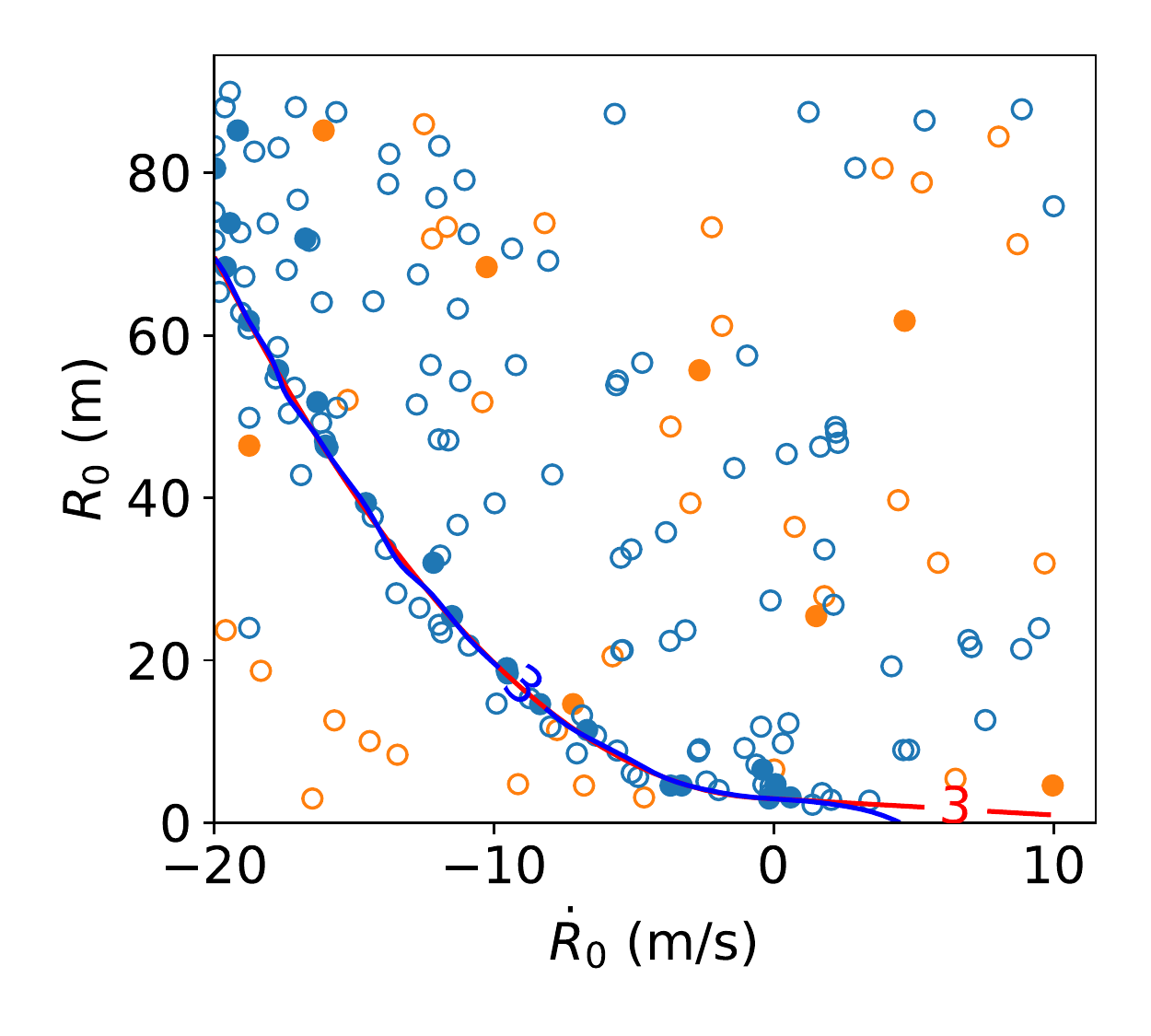}
    \caption{Positions of initial high-fidelity samples (\tikzcircle{2pt, Orange}), initial low-fidelity samples (\tikzcirclehollow{2pt, Orange}),  adaptive high-fidelity samples (\tikzcircle{2pt,NavyBlue}) and adaptive low-fidelity samples (\tikzcirclehollow{2pt, NavyBlue}) from a typical experiment with $f_l$ of $\Delta_t=1$, as well as the learned limiting state (\blueline) compared to the exact one (\redline).}
    \label{fig:cutin_sampling_2}
\end{figure}


Finally, we plot in Fig. \ref{fig:cutin_sampling_2} the positions of the high and low-fidelity (initial and adaptive) samples for a typical experiment of bi-fidelity algorithm for the case with $f_l$ of $\Delta_t=1$. It can be observed that while the adaptive low-fidelity samples are scattered in the space of $\mathbf{x}$, the adaptive high-fidelity samples are most concentrated at the limiting state $\{\mathbf{x}: f_{h,l}(\mathbf{x}) = 3\}$. This is because the limiting state represents the location where the difference between $f_h$ and $f_l$ critically affects the computed value of $P_a$, and it necessarily takes high-fidelity samples to resolve this important region. The resolved limiting state at 60 units of cost in adaptive sampling is also shown in the figure to demonstrate its proximity to the ground truth.

\section{Conclusion}
In this work, we develop an adaptive sampling framework to efficiently evaluate the accident rate $P_a$ of connected and automated vehicles (CAVs) in scenario-based tests. The core components of our approach include a surrogate model by Gaussian process regression and a novel acquisition function to select the next-best sample maximizing its benefit (uncertainty reduction) to $P_a$. The framework can be applied to both single-fidelity and bi-fidelity contexts, where the latter involves a low-fidelity model to help construct the function of the CAV performance. Accordingly, for the latter the two components of our approach need to be extended to the bi-fidelity Gaussian process regression and an acquisition function, allowing the choice of both fidelity level and sampling location. We demonstrate the effectiveness of the framework in a widely-considered two-dimensional cut-in problem, with the low and high-fidelity models constructed by IDM with different resolutions in time. It is shown that the single-fidelity method already outperforms the state-of-the-art method for the same problem, and the bi-fidelity method further accelerates the convergence by a factor of about two (i.e., with half of the computational cost) for most settings that are tested.


{\appendices
\section{Information gain and its approximation}
Let us start from the situation that we have a surrogate model $f_h(\mathbf{x})|\mathcal{D}$ and the associated distribution of $P_a$. From an information-theoretical perspective, the next-best sample $\tilde{\mathbf{x}}$ should be chosen to maximize the information gain for $P_a$, i.e., to maximize the expected KL divergence defined as
\begin{linenomath} \begin{align}
     G(\tilde{\mathbf{x}})
     = & \;  \mathbb{E}_{f_h} \Big(
     \mathrm{KL}\big(p(P_a|\mathcal{D}, \tilde{\mathbf{x}}, f_h(\tilde{\mathbf{x}}))\;||\;p(P_a|\mathcal{D})\big) \Big)
\nonumber \\ 
    = &  \iint p(P_a|\mathcal{D}, \tilde{\mathbf{x}}, f_h(\tilde{\mathbf{x}})) \log \frac{p(P_a|\mathcal{D}, \tilde{\mathbf{x}}, f_h(\tilde{\mathbf{x}}))}{p(P_a|\mathcal{D})} \; \mathrm{d} P_a   
\nonumber \\     
    & \quad * p(f_h(\tilde{\mathbf{x}})|\mathcal{D}) \; \mathrm{d} f_h(\tilde{\mathbf{x}}).
\label{IG_0}
\end{align} \end{linenomath} 
While a direct computation of \eqref{IG_0} is extremely expensive, three assumptions can be made to simplify its expression, which are summarized below. \\

\begin{enumerate}
   \item Instead of a random $f_h(\tilde{\mathbf{x}})$, we assume that it can be approximated by the mean prediction from the current GPR: $\overline{f}_h(\tilde{\mathbf{x}}) =  \mathbb{E}(f_h(\tilde{\mathbf{x}})|\mathcal{D})$.
   Thus \eqref{IG_0} becomes:
\begin{equation}
     G(\tilde{\mathbf{x}}) \approx
       \int p(P_a|\mathcal{D}, \tilde{\mathbf{z}}) \log \frac{p(P_a|\mathcal{D}, \tilde{\mathbf{z}})}{p(P_a|\mathcal{D})} \mathrm{d} P_a,
\label{IG_1}
\end{equation}
where $\tilde{\mathbf{z}} = \{ \tilde{\mathbf{x}},  \overline{f}_h(\tilde{\mathbf{x}})\}$.
   
   \item We assume $P_a$ follows Gaussian distributions with $P_a|\mathcal{D} \sim \mathcal{N}(\mu_1, \sigma^2_1)$ and $P_a|\mathcal{D},\tilde{\mathbf{z}} \sim \mathcal{N}(\mu_2(\tilde{\mathbf{x}}), \sigma^2_2(\tilde{\mathbf{x}}))$. Substitution of these distributions into \eqref{IG_1} gives
   
   \begin{equation}
    G(\tilde{\mathbf{x}}) \approx  \log (\frac{\sigma_1}{\sigma_2(\tilde{\mathbf{x}})}) + \frac{\sigma^2_2(\tilde{\mathbf{x}})}{2 \sigma^2_1} + \frac{(\mu_2(\tilde{\mathbf{x}})-\mu_1)^2}{2 \sigma^2_1} - \frac{1}{2}.
    \label{IG_2} 
    \end{equation}

   \item The difference of $\mu_1$ and $\mu_2$ is much smaller than the standard deviation of $P_a$ i.e. $|\mu_2(\tilde{\mathbf{x}})-\mu_1| \ll \sigma_2(\tilde{\mathbf{x}})$ (which is generally true unless $P_a$ has been estimated very well). This leads to
   \begin{equation}
    G(\tilde{\mathbf{x}}) 
    \approx \log (\frac{\sigma_1}{\sigma_2(\tilde{\mathbf{x}})}) + \frac{\sigma^2_2(\tilde{\mathbf{x}})}{2 \sigma^2_1} - \frac{1}{2}.
    \label{IG_3}
    \end{equation}
\end{enumerate}
It can be shown that $\eqref{IG_3}$ monotonously increases with the decrease of $\sigma_2/ \sigma_1$ in the range of $\sigma_2/\sigma_1<1$. Since a sample always provides information to $P_a$, the condition $\sigma_2/\sigma_1<1$ is always satisfied. Therefore, the maximization of $\eqref{IG_3}$ is equivalent to the minimization of $\sigma_2$, that is consistent the minimization of \eqref{ivr_cov}.

\section{The upper bound of \eqref{ivr_cov}}
Here we show the derivation of the upper bound \eqref{ivr} from \eqref{ivr_cov}:
\begin{linenomath} \begin{align}
    & \mathrm{var}[P_a|\mathcal{D}, \tilde{\mathbf{z}}]
\nonumber \\
    = & \; \mathrm{var}\Big[\int \mathbf{1}_{\delta}\big(f_h(\mathbf{x})|\mathcal{D}, \tilde{\mathbf{z}}\big) p_{\mathbf{x}}(\mathbf{x}){\rm{d}}\mathbf{x}\Big] 
\nonumber \\
    & (\text{hereafter we write} \; \mathbf{1}_{\delta}\big(f_h(\mathbf{x})|\mathcal{D}, \tilde{\mathbf{z}}\big)  \; \text{as} \; \hat{\mathbf{1}}(\mathbf{x}))
\nonumber \\
    = & \; \mathbb{E}\Big[\Big(\int \hat{\mathbf{1}}(\mathbf{x}) p_{\mathbf{x}}(\mathbf{x}){\rm{d}}\mathbf{x}\Big)^2\Big] - \Big(\mathbb{E}\Big[\int \hat{\mathbf{1}}(\mathbf{x}) p_{\mathbf{x}}(\mathbf{x}){\rm{d}}\mathbf{x}\Big] \Big)^2
\nonumber \\ 
    = & \; \mathbb{E} \Big[\int \hat{\mathbf{1}}(\mathbf{x}) p(\mathbf{x}) {\rm{d}}\mathbf{x} \int \hat{\mathbf{1}}(\mathbf{x}') p(\mathbf{x}') {\rm{d}}\mathbf{x}'\Big] 
\nonumber \\ 
     & -  \Big(\mathbb{E} \Big[\int \hat{\mathbf{1}}(\mathbf{x}) p(\mathbf{x}) {\rm{d}}\mathbf{x}\Big]\Big) \Big(\mathbb{E} \Big[\int \hat{\mathbf{1}}(\mathbf{x}') p(\mathbf{x}') {\rm{d}}\mathbf{x}'\Big]\Big)
\nonumber \\ 
    = & \iint \mathbb{E} \Big[\hat{\mathbf{1}}(\mathbf{x})\hat{\mathbf{1}}(\mathbf{x}')\Big] p(\mathbf{x}) p(\mathbf{x}') {\rm{d}}\mathbf{x} {\rm{d}}\mathbf{x}' 
\nonumber \\ 
    & -  \iint \mathbb{E} \Big[\hat{\mathbf{1}}(\mathbf{x})\Big] \mathbb{E} \Big[ \hat{\mathbf{1}}(\mathbf{x}')\Big] p(\mathbf{x}) p(\mathbf{x}') {\rm{d}}\mathbf{x} {\rm{d}}\mathbf{x}' 
\nonumber\\ 
    = & \iint {\rm{cov}} \big[\hat{\mathbf{1}}(\mathbf{x}), \hat{\mathbf{1}}(\mathbf{x}')\big] p(\mathbf{x}) p(\mathbf{x}') {\rm{d}}\mathbf{x} d\mathbf{x}'  
\label{bound0} \\ 
    = & \int {\rm{std}} \big[\hat{\mathbf{1}}(\mathbf{x})\big]  \Big( \int {\rm{std}} \big[\hat{\mathbf{1}}(\mathbf{x}')\big] \rho  \big[ \hat{\mathbf{1}}(\mathbf{x}),\hat{\mathbf{1}}(\mathbf{x}')\big] p(\mathbf{x}') d\mathbf{x}' \Big) p(\mathbf{x})  {\rm{d}}\mathbf{x}
\nonumber \\ 
    \leq & \; 0.5 \int {\rm{std}} \big[\hat{\mathbf{1}}(\mathbf{x})\big] p(\mathbf{x}) {\rm{d}}\mathbf{x},
\label{bound}
\end{align} \end{linenomath} 
where $\rho[\cdot,\cdot]$ denotes the correlation coefficient. The last inequality comes from  ${\rm{std}} \big[\hat{\mathbf{1}}(\mathbf{x})\big] \leq 0.5$  and  $\rho  \big[ \hat{\mathbf{1}}(\mathbf{x}),\hat{\mathbf{1}}(\mathbf{x}')\big] \leq 1$. The former inequality is due to that $\hat{\mathbf{1}}(\mathbf{x})$ is a Bernoulli variable whose standard deviation is maximized to be 0.5 when the two branches $\hat{\mathbf{1}}(\mathbf{x})= 0$ and $\hat{\mathbf{1}}(\mathbf{x})= 1$ both have probability of 0.5.

We can also show that \eqref{ivr_cov} can be directly approximated as $ \sim \int {\rm{var}} \big[\hat{\mathbf{1}}(\mathbf{x})\big] p(\mathbf{x}) {\rm{d}}\mathbf{x}$ if only  variance terms are kept in \eqref{bound0}. This form is somewhat similar to the upper-bound result \eqref{bound} when used as an acquisition function. We apply the upper bound \eqref{bound} in the current paper.

\section{Benchmark tests of the framework}
In this section, we test our (single-fidelity) framework in two benchmark problems to demonstrate the efficiency of our new acquisition function. In each problem, we report the median value as well as 15\% and 85\% percentiles of the exceeding probability (defined below) obtained from 100 applications of our method with different initial samples. 

The first example we consider is a multi-modal function (see Fig. \ref{fig:mm1} for a contour plot) that is also studied in \cite{hu2016global} and \cite{yi2021active}:
\begin{linenomath} \begin{align}
f_h(x_1, x_2)  = & \;\frac{((1.5 + x_1)^2+4)(1.5 + x_2)}{20}  - \sin(\frac{7.5 + 5x_1}{2})
\nonumber \\
                 & \;  - 2.
\end{align} \end{linenomath} 
The input $\mathbf{x}$ follows a Gaussian distribution $p(x_1, x_2) = \mathcal{N}(\boldsymbol{0}, \mathrm{I})$, with $\mathrm{I}$ being a $2 \times 2$ identity matrix. We are interested in the exceeding probability $P_e = \int \mathbf{1}_{f_h>0}(f_h(\mathbf{x}))p_{\mathbf{x}}(\mathbf{x}){\rm{d}}\mathbf{x}$. In the application of our method, we use 8 random initial samples followed by 22 adaptive samples, with the computed $P_e$ plotted in Fig. \ref{fig:mm2} as a function of the sample numbers. Also shown in Fig. \ref{fig:mm2} is the 3\% error bounds of ground truth. We see that percentiles of $P_e$ converge to the error bounds in 18 samples. The sampling positions of our method are plotted in Fig. \ref{fig:mm3} as well as the computed limiting state after 30 samples with comparison to the ground truth.

Other existing approaches were also tested for the same problem for the number of samples leading to convergence, but usually with less strict (or different) criterion defined for convergence. The number of samples are 19 \cite{hu2016global}, 31$-$44 \cite{ echard2011ak}, 36$-$69 \cite{bichon2008efficient}. 

\begin{figure}
    \centering
    \includegraphics[width=7cm]{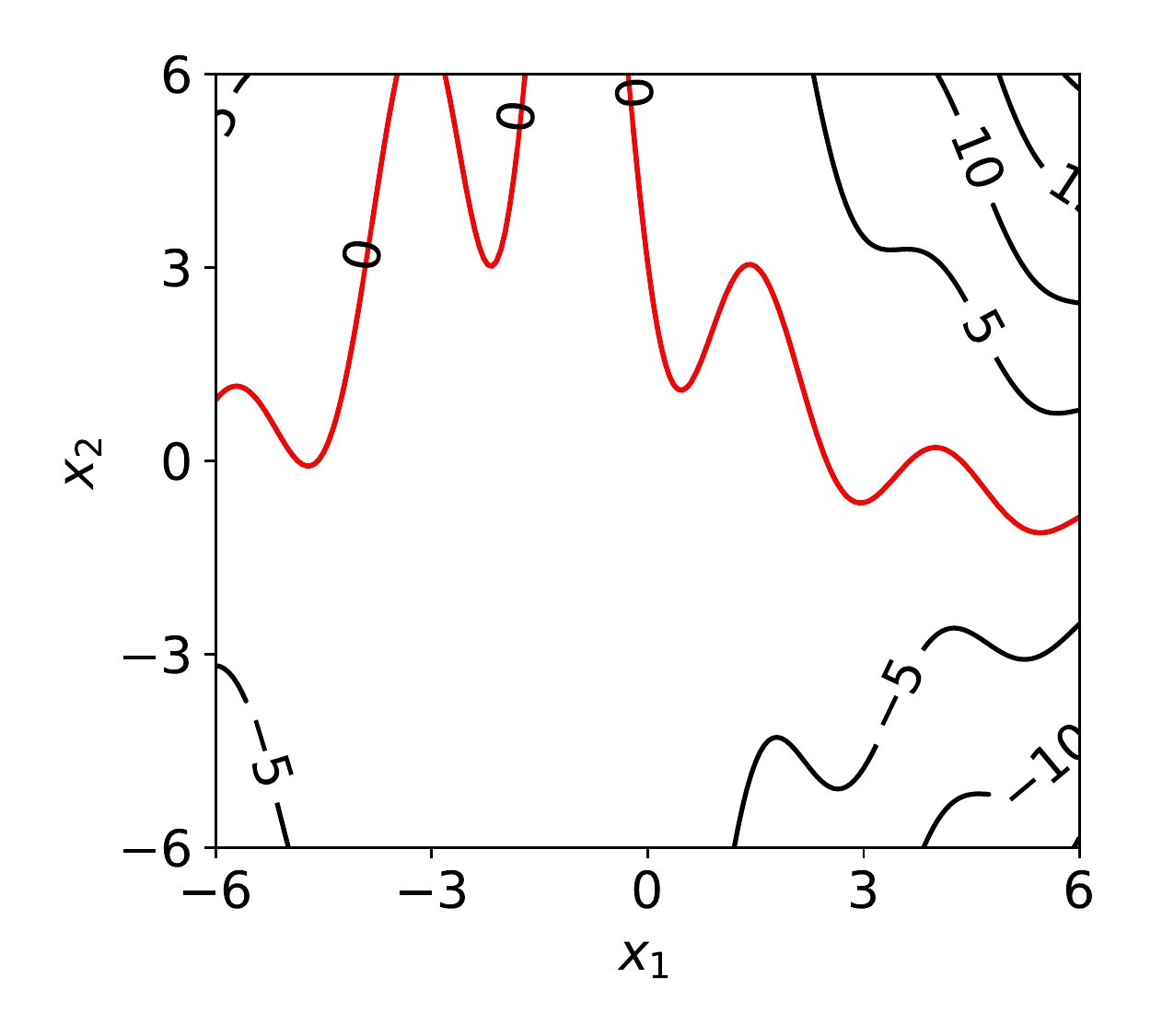}
    \caption{$f_h(x_1, x_2)$ of the multi-modal function with the limiting state $\{\mathbf{x}: f_{h}(x_1, x_2) = 0\}$  (\redline).}
    \label{fig:mm1}
\end{figure}

\begin{figure}
    \centering
    \includegraphics[width=7.5cm]{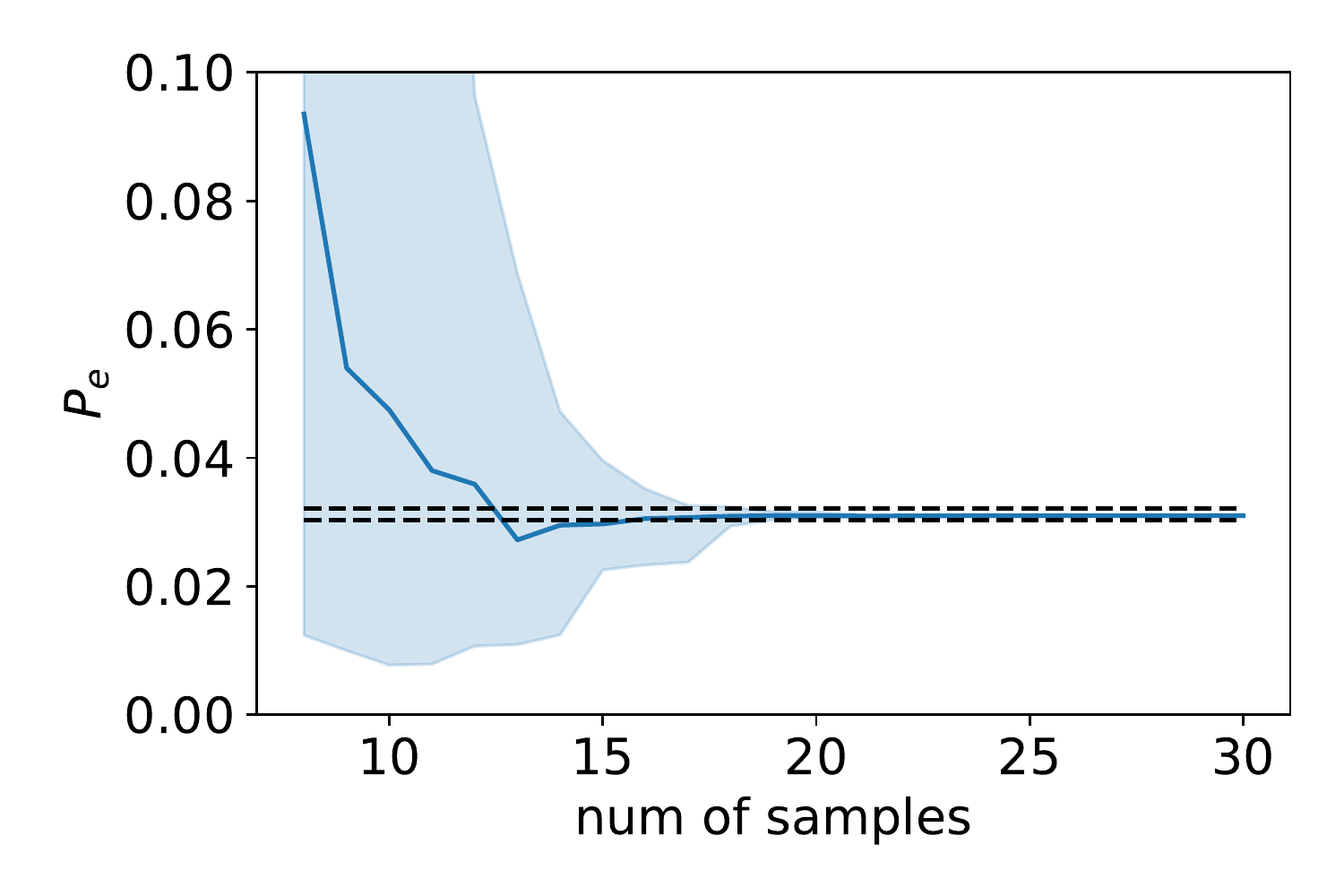}
    \caption{Results of single-fidelity method for the problem of multi-modal function, presented by the median value (\blueline) as well as the 15\% and 85\% percentiles (shaded region) from 100 experiments. The ground-truth of $P_e$ is shown (\blackdashedline) in terms of the 3\% error bounds. }
    \label{fig:mm2}
\end{figure}

\begin{figure}
    \centering
    \includegraphics[width=7cm]{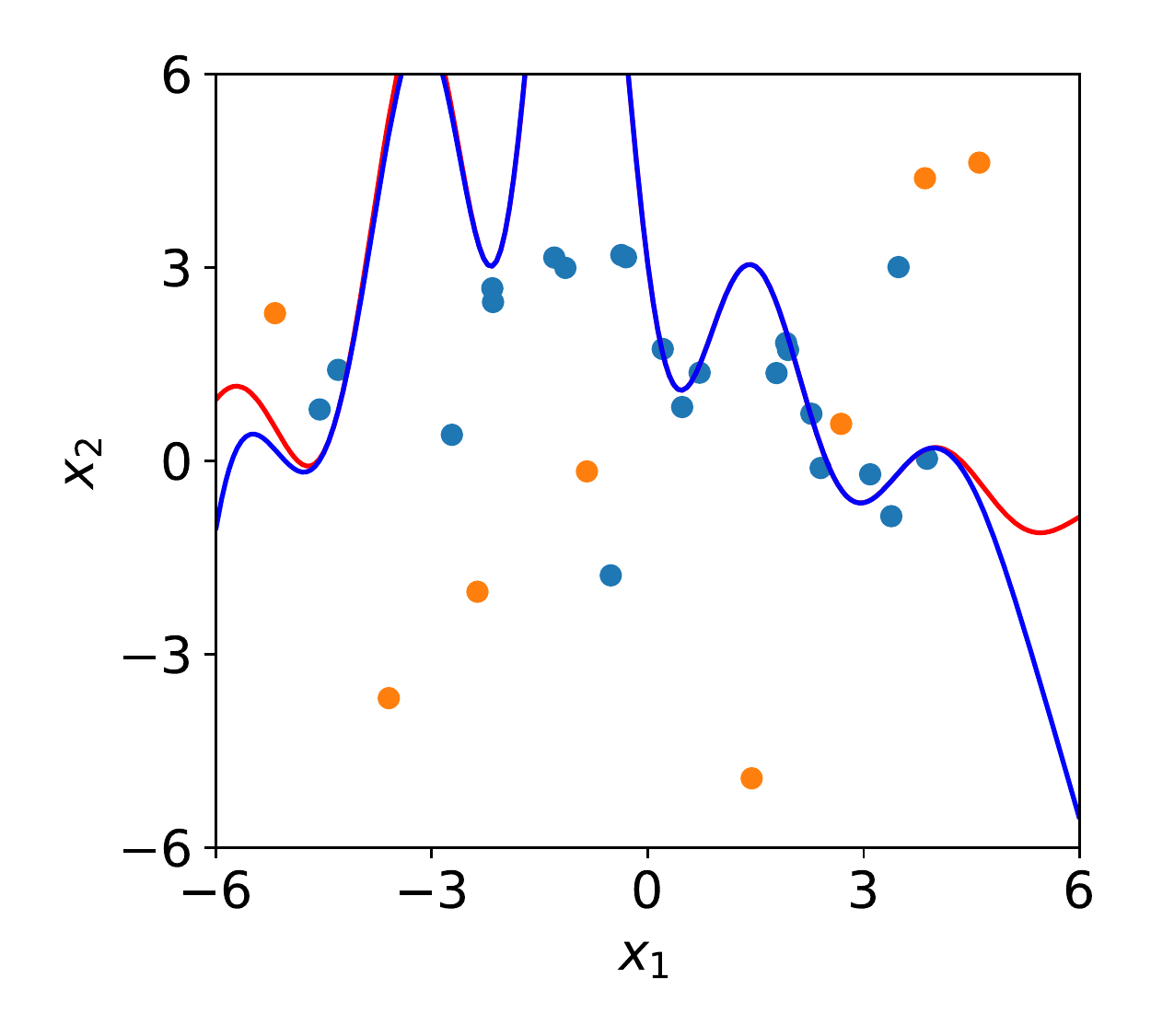}
    \caption{Typical positions of 8 initial samples (\tikzcircle{2pt, Orange}) and 22 adaptive samples (\tikzcircle{2pt,NavyBlue}) for the problem of multi-modal function, as well as the learned limiting state (\blueline) compared to the exact one (\redline).}
    \label{fig:mm3}
\end{figure}

The second example is the four-branch function (Fig. \ref{fig:fb1}):
$$
f(x_1, x_2) = - {\rm{min}}\left\{
\begin{aligned}
    & 3 + 0.1(x_1-x_2)^2 + \frac{(x_1 + x_2)}{\sqrt{2}}                     \\
    & 3 + 0.1(x_1-x_2)^2 - \frac{(x_1 + x_2)}{\sqrt{2}}                       \\
    & (x_1 - x_2) + \frac{6}{\sqrt{2}}           \\
    & (x_2 - x_1) + \frac{6}{\sqrt{2}}           \\
\end{aligned}
\right..
$$
The input $\mathbf{x}$ follows a Gaussian distribution $p(x_1, x_2) = \mathcal{N}(\boldsymbol{0}, \mathrm{I})$. Our method is applied with 12 random initial samples followed by 68 adaptive samples, with $P_e$ plotted in Fig. \ref{fig:fb2} as a function of sample numbers. Also shown in Fig. \ref{fig:fb2} is the 3\% error bounds of ground truth. We see that percentiles of $P_e$ converge to the error bounds in 42 samples. The sampling positions of our method are plotted in Fig. \ref{fig:fb3} as well as the computed limiting state after 80 samples with comparison to the ground truth. 

With less strict (or different) convergence criterion, the number of samples leading to the convergence in existing works are 36 \cite{hu2016global}, 38 \cite{sun2017lif}, 78 \cite{schobi2015polynomial}, 65$-$126 \cite{echard2011ak},  68$-$124 \cite{bichon2008efficient}, and 167 \cite{marelli2018active}.

\begin{figure}
    \centering
    \includegraphics[width=7cm]{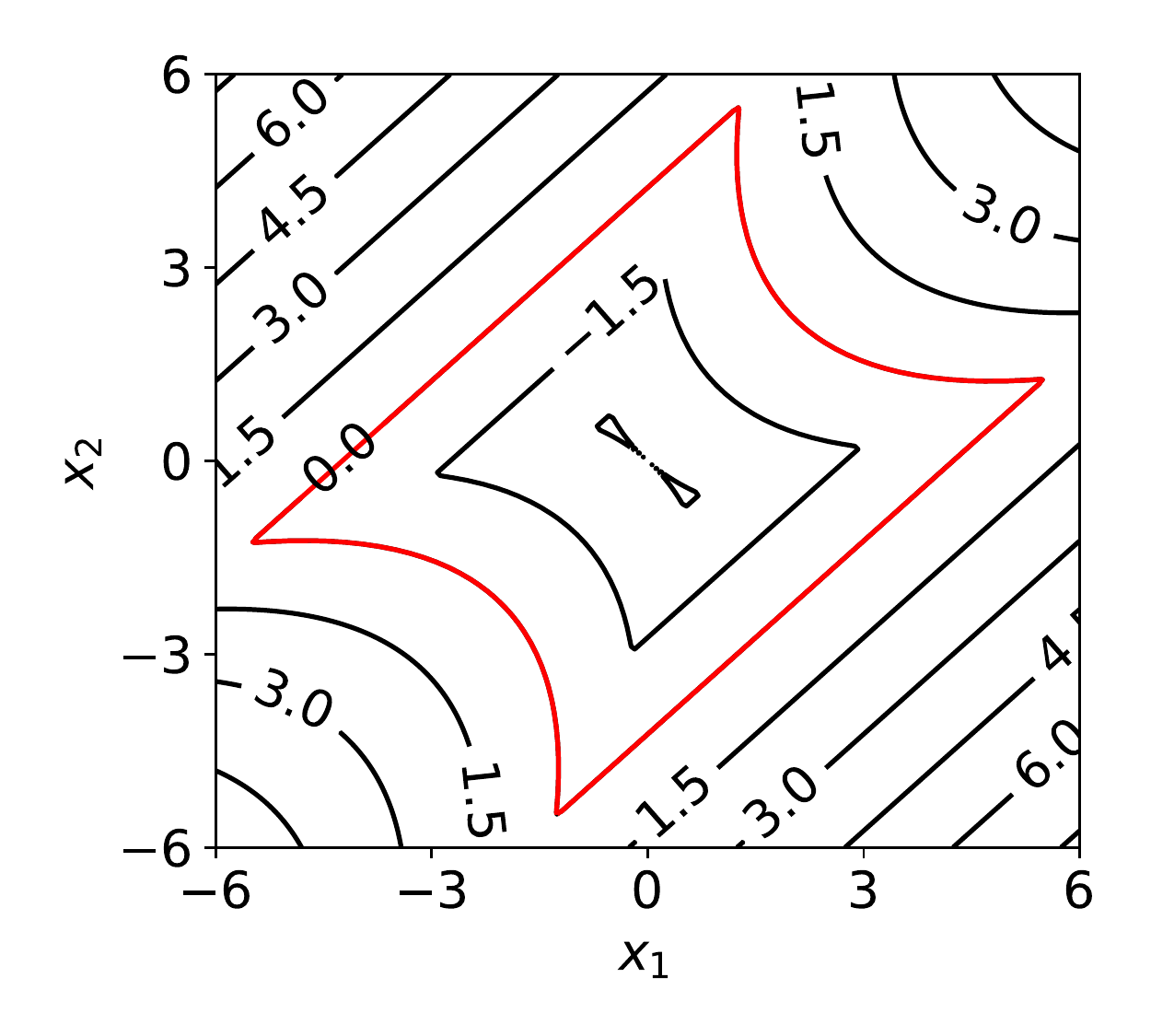}
    \caption{$f_h(x_1, x_2)$ of the four-branch function with the limiting state $\{\mathbf{x}: f_{h}(x_1, x_2) = 0\}$  (\redline).}
    \label{fig:fb1}
\end{figure}

\begin{figure}
    \centering
    \includegraphics[width=7.5cm]{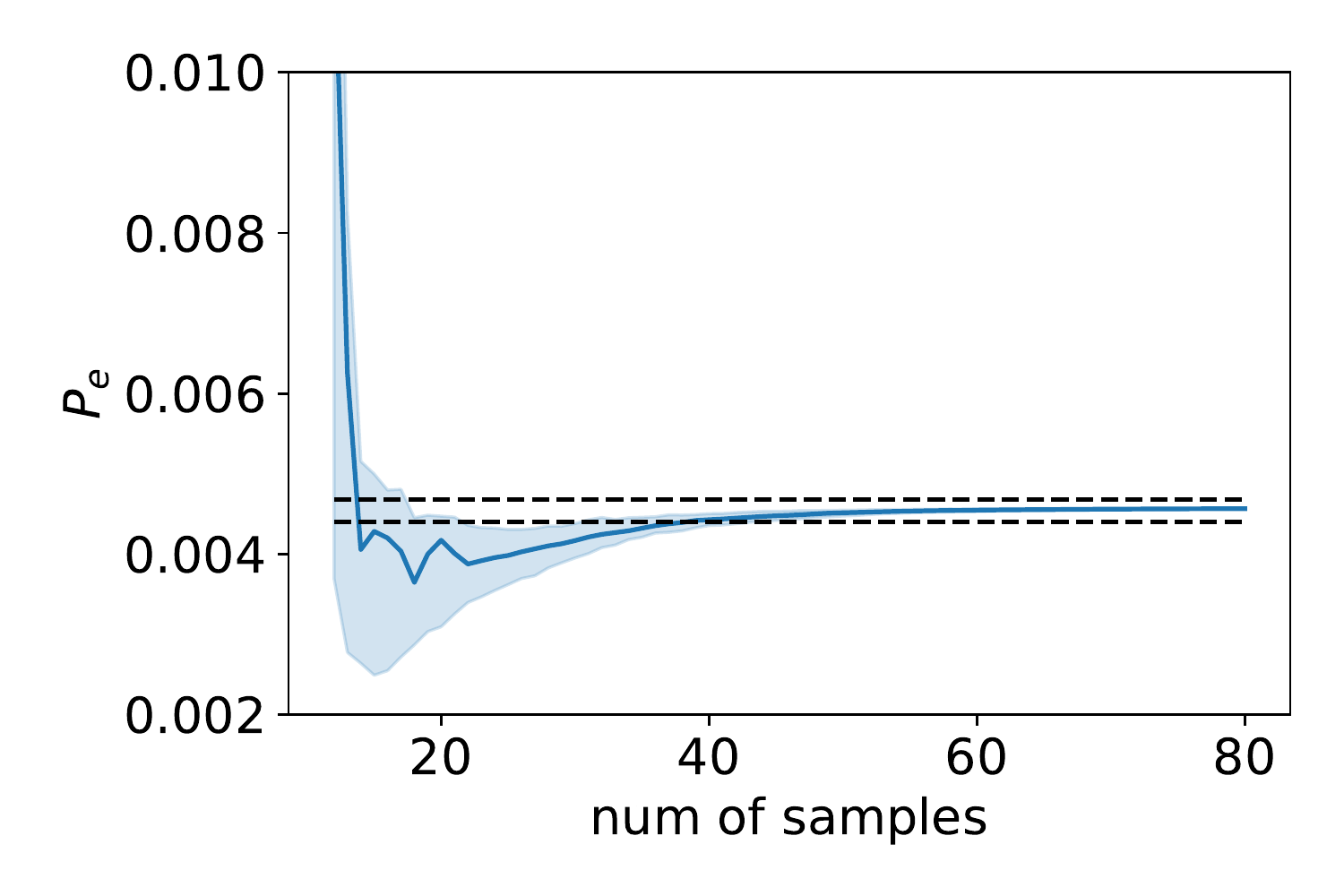}
    \caption{Results of single-fidelity method for the problem of four-branch function, presented by the median value (\blueline) as well as the 15\% and 85\% percentiles (shaded region) from 100 experiments. The ground-truth of $P_e$ is shown (\blackdashedline) in terms of the 3\% error bounds.}
    \label{fig:fb2}
\end{figure}
    
\begin{figure}
    \centering
    \includegraphics[width=7cm]{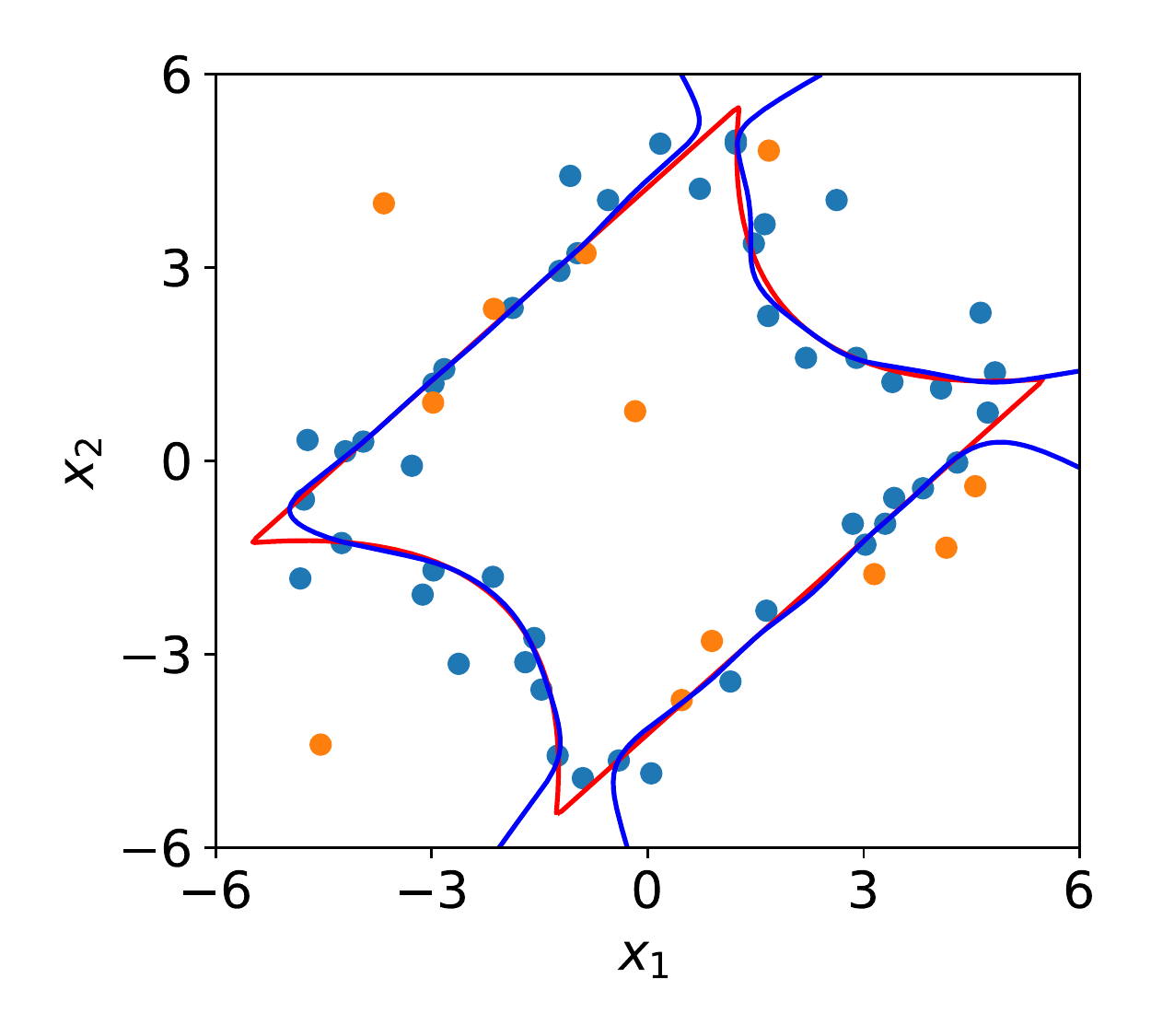}
    \caption{Typical positions of 12 initial samples (\tikzcircle{2pt, Orange}) and 68 adaptive samples (\tikzcircle{2pt,NavyBlue}) for the problem of four-branch function, as well as the learned limiting state (\blueline) compared to the exact one (\redline).}
    \label{fig:fb3}
\end{figure}
}

\bibliographystyle{IEEEtran}
\bibliography{reference.bib}

\newpage

\begin{IEEEbiography}
[{\includegraphics[width=1in,height=1.25in,clip,keepaspectratio]{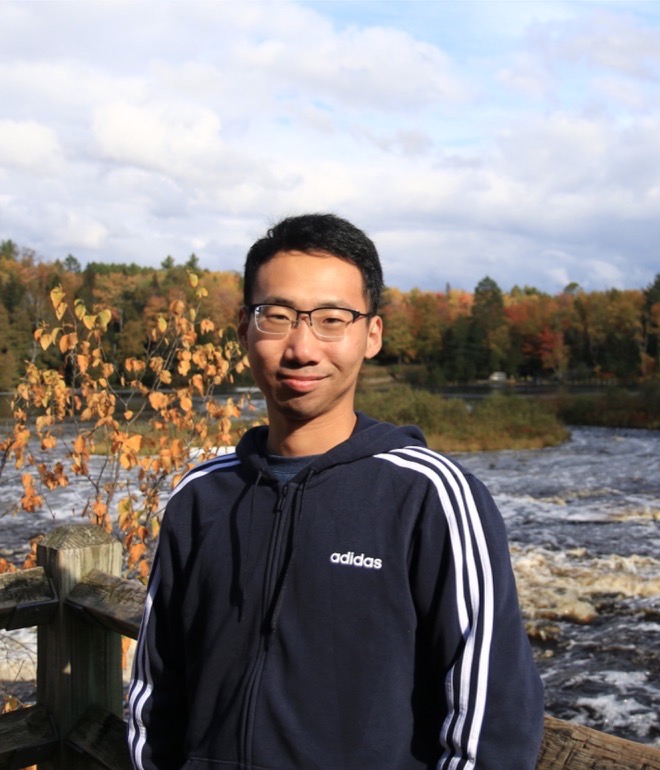}}]{Xianliang Gong} received a bachelor’s degree from Harbin Engineering University in 2015 and master’s degree from Shanghai Jiao Tong University in 2018. He is currently a Ph.D. student at the department of naval architecture and marine engineering, University of Michigan. His research interests are uncertainty quantification and prediction of extreme events by machine learning and physical modeling. 
\end{IEEEbiography}

\begin{IEEEbiography}
[{\includegraphics[width=1in,height=1.25in,clip,keepaspectratio]{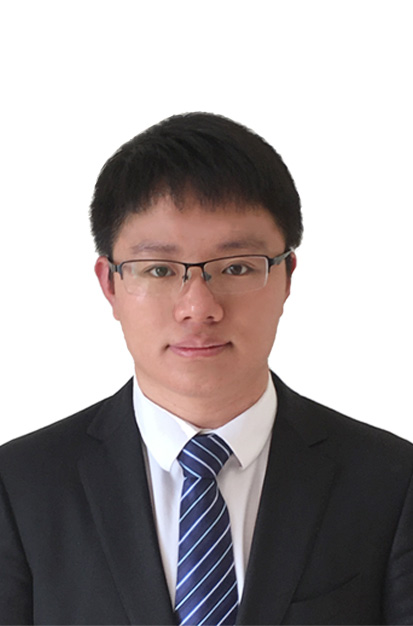}}]{Shuo Feng} (Member, IEEE) received bachelor’s and Ph.D. degrees in the Department of Automation at Tsinghua University, China, in 2014 and 2019, respectively. He was also a joint Ph.D. student from 2017 to 2019 and a postdoctoral research fellow from 2019 to 2021 in the Department of Civil and Environmental Engineering at the University of Michigan, Ann Arbor. He is currently an Assistant Research Scientist at the University of Michigan Transportation Research Institute (UMTRI). His research interests lie in the development and validation of safety-critical machine learning, particularly for connected and automated vehicles. He is the Associate Editor of the \textit{IEEE Transactions on Intelligent Vehicles} and the Academic Editor of the \textit{Automotive Innovation}. He was the recipient of the Best Ph.D. Dissertation Award from the IEEE Intelligent Transportation Systems Society in 2020 and the ITS Best Paper Award from the INFORMS TSL Society in 2021.
\end{IEEEbiography}

\begin{IEEEbiography}
[{\includegraphics[width=1in,height=1.25in,clip,keepaspectratio]{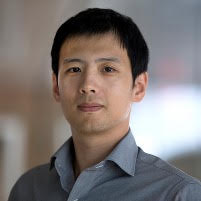}}]{Yulin Pan} received his bachelor's degree from Huazhong University of Science and Technology, master's degree from the University of Texas at Austin and Ph.D. from Massachusetts Institute of Technology, respectively from ocean/civil/mechanical engineering. He is now an assistant professor in the department of naval architecture and marine engineering at the University of Michigan, Ann Arbor. His research interests include theoretical and computational hydrodynamics, nonlinear waves and turbulence theory, applied mathematics, and statistical reliability analysis.  
\end{IEEEbiography}

\vfill

\end{document}